\documentclass[preprint,1p,times]{elsarticle}

\usepackage[utf8]{inputenc}
\usepackage[dvipsnames,table,xcdraw]{xcolor}
\usepackage{amsmath,amsfonts,bbm}
\usepackage{floatrow,subcaption}
\usepackage{hyperref}
\usepackage{tikz}
\usetikzlibrary{shapes,arrows}

\tikzstyle{decision} = [diamond, draw, fill=green!40, 
    text width=4.5em, text badly centered, node distance=3cm, inner sep=0pt]
\tikzstyle{block} = [rectangle, draw, fill=cyan!40, 
    text width=5em, text centered, rounded corners, minimum height=4em]
\tikzstyle{special block} = [rectangle, draw, fill=cyan, 
    text width=5em, text centered, rounded corners, minimum height=4em]
\tikzstyle{line} = [draw, -latex']
\tikzstyle{cloud} = [draw, ellipse, fill=gray!40, node distance=3cm,
    minimum height=2em]

\newfloatcommand{capbtabbox}{table}[][0.48\textwidth]

\begin{document}

\begin{frontmatter}

\title{E2N: Error Estimation Networks for Goal-Oriented Mesh Adaptation}

\author[1]{Joseph G. Wallwork\corref{cor}}
\ead{j.wallwork16@imperial.ac.uk}

\author[1,2]{Jingyi Lu}

\author[1]{Mingrui Zhang}

\author[1]{Matthew D. Piggott}

\cortext[cor]{Corresponding author}

\address[1]{Department of Earth Science and Engineering, Imperial College London, London, SW7 2AZ, UK}
\address[2]{Huawei Technologies Co., Ltd, Shenzhen, China (current address)}

\journal{Journal of Computational Physics}

\begin{abstract}
    Given a partial differential equation (PDE), goal-oriented error estimation allows us to understand how errors in a diagnostic quantity of interest (QoI), or \emph{goal}, occur and accumulate in a numerical approximation, for example using the finite element method.
    By decomposing the error estimates into contributions from individual elements, it is possible to formulate adaptation methods, which modify the mesh with the objective of minimising the resulting QoI error.
    However, the standard error estimate formulation involves the true adjoint solution, which is unknown in practice.
    As such, it is common practice to approximate it with an `enriched' approximation (e.g.~in a higher order space or on a refined mesh).
    Doing so generally results in a significant increase in computational cost, which can be a bottleneck compromising the competitiveness of (goal-oriented) adaptive simulations.
    The central idea of this paper is to develop a ``data-driven'' goal-oriented mesh adaptation approach through the selective replacement of the expensive error estimation step with an appropriately configured and trained neural network.
    In doing so, the error estimator may be obtained without even constructing the enriched spaces.
    An element-by-element construction is employed here, whereby local values of various parameters related to the mesh geometry and underlying problem physics are taken as inputs, and the corresponding contribution to the error estimator is taken as output.
    We demonstrate that this approach is able to obtain the same accuracy with a reduced computational cost, for adaptive mesh test cases related to flow around tidal turbines, which interact via their downstream wakes, and where the overall power output of the farm is taken as the QoI.
    Moreover, we demonstrate that the element-by-element approach implies reasonably low training costs.
\end{abstract}


\end{frontmatter}

\section{Introduction}\label{sec:intro}

\subsection{Goal-Oriented Error Estimation}\label{subsec:intro:go}

Numerical simulations across the majority of science and engineering often have an associated diagnostic \emph{quantity of interest (QoI)}.
This could be the drag or lift on an aeroplane wing, or the level of radiation in a particularly sensitive region in the vicinity of a nuclear power plant, for example.
As indicated in the abstract, the motivating application utilised in this paper is in the context of tidal resource assessment for the marine renewable energy sector, whereby the instantaneous power or total energy output of a tidal farm provide obvious choices for QoIs \cite{PK22}.
The abstract and idealised nature at which this is considered makes the results highly relevant to the closely related wind turbine application.

\emph{Goal-oriented error estimation} provides a framework for assessing the error in the chosen QoI due to a particular numerical approximation method for solving the underlying PDE.
This is achieved by means of the \emph{adjoint problem} associated with the PDE and the QoI.
Goal-oriented error estimates are typically derived from the \emph{dual weighted residual (DWR)} \cite{BR01}.
The `first order' DWR result involves the weak residual of the PDE -- assuming some finite element discretisation -- but with the test function replaced with the solution of the adjoint problem.
Since this `exact' adjoint solution is not obtainable in practice, we are unable to evaluate the error estimate without the use of further approximation methods.

One option is to apply error estimation techniques, such as a posteriori upper bounds, to construct an alternative formulation, wherein the true adjoint solution does not appear
(see, for example, the `difference quotient' approach in \cite{BR01}).
Another approach is to approximate the adjoint solution in an enriched finite element space.
This can be done by solving the PDE in a globally enriched space (see \cite{WB20}, for example), solving PDEs in locally enriched spaces (see \cite{DR17}, for example), or by applying superconvergent patch recovery techniques (see \cite{ZZ87}, for example).
In any case, the enrichment procedure tends to be computationally expensive, although the level of expense can vary between approaches.

\subsection{Accelerating Mesh Adaptation using Neural Networks}\label{subsec:intro:ann}

Goal-oriented error estimation is most commonly deployed for the purposes of mesh adaptation.
By determining local contributions to an estimator (i.e.~\emph{`error indicators'}), it is possible to establish parts of the mesh that would benefit from increased resolution, as well as parts of the mesh with unnecessary degrees of freedom (DoFs).
The resulting numerical accuracy versus efficiency trade-off may be interpreted as a constrained optimisation problem, whereby we seek to minimise QoI error, subject to having a mesh with (approximately) some target number of DoFs.
However, for the reasons mentioned above, the goal-oriented error estimation step typically comes at significant computational expense.
In many cases, it can be more computationally demanding than solving the underlying prognostic PDE, making it a bottleneck within the goal-oriented mesh adaptation workflow.

Neural networks provide an exciting new way to accelerate advanced discretisation methods like goal-oriented mesh adaptation.
Given the computational expense of such methods, there is potential for massive gains in efficiency and widening of access to users that would otherwise view them as too time intensive.
As remarked in \cite[p.15]{HK21}, ``[the application of neural networks to emulate mesh adaptation] in the general CFD pipeline is one of the most risk-averse ways of using machine learning in this context''.
That is, the worst that can happen is that the resulting adapted mesh is of relatively poor quality and the solver convergence suffers as a consequence.
On the other hand, if the PDE solve step is to be emulated, for example, then an incorrect application could potentially result in unrealistic simulation data, i.e.~``incorrect physics''.

Whilst this is a relatively new field of research, there are already a number of existing works.
A variety of different approaches have been taken, with different authors advocating the replacement of different parts of the (goal-oriented) mesh adaptation workflow by neural networks.
These include the test function choice \cite{BM21}, forward solve \cite{WL21}, adjoint solve \cite{RS21}, derivative recovery procedure \cite{KM20}, error estimation \cite{CW21}, metric/monitor function/sizing field construction step \cite{PF20,FC21,CF21}, and the entire mesh adaptation loop \cite{HK21,YD21,SZ22}.


One of the most relevant references for this work is \cite{FC21}.
In that work, a neural network is used to `solve' the optimisation problem used to determine a metric tensor from solution information in the \emph{Mesh Optimization through
Error Sampling and Synthesis (MOESS)} approach \cite{YD12}.
Each iteration of the MOESS algorithm requires the consideration of all possible modifications that can be made to each mesh element and the solution of local PDEs defined on patches of modified elements and their neighbours.
The application of neural networks means that such sub-meshes and sub-problems do not need to be explicitly considered, as the mapping from the current state to the optimised Riemannian metric is `learnt' implicitly during training.
This methodology is tested on steady-state `flow past an airfoil' test cases in \cite{FC21} and is extended to use convolutional neural networks in \cite{CF21}.

The authors of \cite{HK21} point out that, whilst goal-oriented mesh adaptation is a `gold-standard' approach to mesh adaptation, it can be prohibitively computationally expensive and even infeasible to use, given that many solvers do not have automated adjoint capability.
That work goes further than emulating the metric construction step of the mesh adaptation pipeline, covering the forward and adjoint PDE solve steps as well, so that the user of a trained network does not need to solve adjoint problems at all.
Goal-oriented mesh adaptation is applied to an unsteady compressible flow solver in a range of rectangular domains with obstacles with randomly generated geometries, where the QoI is the drag on the object.
The domain is interpreted as a grayscale image comprised of pixels, which may be interpreted as a rectangular input array so that a deep convolutional neural network can be applied.
The output is a mesh sizing field, which can be likened to a monitor function or an isotropic metric field.


The works discussed in the previous two paragraphs concern topology-deforming (i.e.~$h$-adaptive) methods.
There also exist several papers concerning topology-preserving (i.e.~$r$-adaptive) methods, although none of these methods use the goal-oriented approach.
One example is \cite{WL21}, which uses the variational mesh movement framework of \cite{HK15}.
In that framework, the metric tensor now prescribes an anisotropic extension of the `mesh density' concept.
Given the current mesh at a timestep, a metric tensor is constructed based on the associated PDE solution, often in terms of its gradient or curvature.
A new mesh that is optimal w.r.t.~the metric tensor is determined by the iterative solution of a so-called \emph{mesh movement partial differential equation (MMPDE)} \cite{BH09}.
MMPDEs are usually nonlinear and are therefore solved numerically using iterative methods.
At each iteration, the metric must be recomputed.
However, it is undesirable to solve the physical PDE multiple times during this process from a computational efficiency perspective.
As such, the authors of \cite{WL21} train a neural network to emulate the PDE solution itself and use this approximate approach at each iteration of the MMPDE solve.
They then solve the full PDE using conventional methods on the final mesh, so that physics are accurately represented.
The authors of \cite{SZ22} go further and propose an end-to-end $r$-adaptation method, whereby the solution on the current mesh is mapped directly to the adapted mesh, removing the need to solve any MMPDEs.
This can be particularly beneficial for highly nonlinear MMPDEs such as those of Monge-Amp\`ere type, which can be computationally expensive to solve.

As well as through the use of a remesher, there exist methods for optimising the input mesh directly.
In \cite{BE20}, an adjoint-based differentiable fluid simulator is embedded into a graph convolution network based model as an additional layer.
With the gradient information computed by the adjoint method, the input mesh of the simulator is regarded as a trainable parameter, to be trained jointly with the network.
Similarly, \cite{ZY20} and \cite{NP21} use neural networks to guide mesh generation.
From the user perspective, \cite{HK21} also describes a mesh generator, albeit one that emulates a goal-oriented mesh adaptation loop.

As mentioned above, \cite{CW21} emulates goal-oriented error estimation using neural networks.
However, that work uses a collocation method based on B-splines to discretise the strong form of a PDE, as opposed to discretising a weak form using finite elements.
As such, technically it does not do mesh adaptation, since there is no `mesh' concept.
However, the fact that collocation point locations are optimised means that the discretisation is being adapted based on the data-driven error estimation process.

\subsection{Novel Aspects of the Present Work}\label{subsec:intro:novel}

Previous work by the authors of this paper has applied goal-oriented error estimation techniques to tidal energy resource assessment with power or energy output as the QoI (see \cite{WB20} for the steady-state case and \cite{WA22} for the time-dependent case).
We build upon this work and seek to accelerate the numerical experiments presented in \cite{WB20} using neural networks.
For consistency with that work, we opt to compute the enriched adjoint solution by solving for it on a globally uniformly refined mesh.
For triangular meshes in 2D, this implies an auxiliary PDE solve on a mesh with four times as many elements as the base mesh (sometimes referred to as `red' refinement), in which case the error estimation step becomes a significant portion of the overall computational cost.

Recall that the authors of \cite{FC21} train a neural network to emulate a metric construction strategy.
In this work, we instead train a neural network to emulate the error estimation strategy.
This choice is made for a number of reasons.
Firstly, it means that we retain the flexibility to use any metric formulation -- rather than being constrained to the one that the network was trained on -- or indeed to use a different ``optimal mesh'' concept entirely, such as one using a monitor function.
In a similar vein, we do not need to use the output of the network for mesh adaptation at all and could perform error estimation analysis instead.
Finally, we seek to apply the neural network in a focused or `surgical' manner, to a single, computationally expensive component of the mesh adaptation workflow, as opposed to treating it as a `black box' mesh adaptation algorithm.

A similar `focused' approach is also used in \cite{RS21}, which emulates the adjoint solve procedure.
This is done on the base mesh and the data-driven adjoint solution is projected into an enriched space, where error indicators are assembled and thereby used to drive mesh adaptation.
The main difference herein is that we avoid the construction of enriched finite element spaces altogether, using a neural network to map directly from objects defined on the base mesh to an error indicator on that same mesh.
By avoiding the construction of enriched spaces, our approach reduces the memory footprint of the resulting algorithm significantly.

The remainder of the paper is as follows.
Section \ref{sec:background} provides background material on goal-oriented error estimation and mesh adaptation, as well as introducing the method used for tidal turbine modelling.
Section \ref{sec:meth} describes the methodology for emulating the goal-oriented error estimation process, which is applied to steady-state tidal turbine test cases in Section \ref{sec:numexp}.
Finally, conclusions are drawn and an outlook on future work is made in Section \ref{sec:conc}.
Details on the derivative recovery procedure and software used to generate the numerical results presented in this paper are provided in \ref{sec:recovery} and \ref{sec:software}.

\section{Background}\label{sec:background}

\subsection{Forward and Adjoint Problems}\label{subsec:background:problems}

Consider a PDE with solution $u$ from a function space $V$ and a variational formulation,
\begin{equation}\label{eq:background:var_form}
    \rho(u,v)=0,\quad\forall v\in V,
\end{equation}
where the residual $\rho:V\times V\rightarrow\mathbb R$ is linear in its second argument.
For the purposes of this paper, we consider steady-state problems, meaning that all derivatives are spatial and there is no time-dependence.

Suppose there exists a diagnostic QoI, $J:V\rightarrow\mathbb R$, which we would like to accurately evaluate.
Given such a QoI, we may formulate the adjoint equation associated with (\ref{eq:background:var_form}) as
\begin{equation}\label{eq:background:adj_var_form}
    \mathcal L_u\:\rho(v,u^*)=\mathcal L_u\:J(v),\quad\forall v\in V,
\end{equation}
where $\mathcal L_w\:\rho:V\times V\rightarrow\mathbb R$ and $\mathcal L_w\:J:V\rightarrow\mathbb R$ denote the linearisations of $\rho$ and $J$ about their first arguments, in direction $w\in V$.
Here $u^*$ is the so-called \emph{adjoint solution}, which (\ref{eq:background:adj_var_form}) is linear with respect to.

In the finite element method, we approximate $u$ in a finite-dimensional subspace, $V_h\subset V$, based upon a mesh $\mathcal H_h$ of the spatial domain.
This gives rise to a weak formulation for (\ref{eq:background:var_form}),
\begin{equation}\label{eq:background:weak_form}
    \rho(u_h,v)=0,\quad\forall v\in V_h,
\end{equation}
where $u_h\in V_h\subset V$ is the finite element approximation.
Similarly, we approximate the adjoint solution by solving the discrete adjoint equation
\begin{equation}\label{eq:background:adj_weak_form}
    \mathcal L_{u_h}\:\rho(v,u^*_h)=\mathcal L_{u_h}\:J(v),\quad\forall v\in V_h.
\end{equation}
Note that the finite element adjoint problem is linearised about $u_h$, rather than $u$.

\subsection{Error Estimation and Indication}\label{subsec:background:ee}

We are primarily interested in goal-oriented error estimators of \emph{dual weighted residual (DWR)} type.
In particular, the `first order' DWR estimator, $\mathcal E$, which satisfies the relation
\begin{equation}\label{eq:background:dwr}
    J(u)-J(u_h)=\underbrace{\rho(u_h,u^*-u^*_h)}_{\mathcal E}+R^{(2)},
\end{equation}
where the remainder term $R^{(2)}$ is quadratic in the forward and adjoint errors $u-u_h$ and $u^*-u^*_h$.
Such an error estimator can be decomposed on an element-wise basis, in order to convey the error contribution due to each mesh element:
\begin{equation}\label{eq:background:estimator}
    \mathcal E=\sum_{K\in\mathcal H_h}\mathcal E_K,\qquad
    \mathcal E_K:=\rho(u_h,u^*-u^*_h)|_K.
\end{equation}
This decomposition can be used to generate an indicator field, $i_h$, which takes a constant value on each element:
\begin{equation}\label{eq:background:indicator}
    i_h:=\sum_{K\in\mathcal H_h}i_K,\qquad
    i_K(\mathbf x):=\left\{\begin{array}{cc}
        \mathcal E_K & \mathbf x\in K \\
        0 & \mathbf x\not\in K
    \end{array}\right..
\end{equation}

The DWR estimator provides a first order approximation to the error in the QoI by combining residuals of the forward equation and sensitivity information related to the QoI via the adjoint solution.
However, as discussed in Section \ref{sec:intro}, there is a practical issue: the estimator involves the true adjoint solution, $u^*$, which is in general unknown.
As such, it is not possible to use $\mathcal E$ in the form presented in (\ref{eq:background:dwr}) and we instead use an approximate `enrichment method'.
The main research question in this paper is: \emph{can we reliably emulate the (typically) expensive procedure using efficient machine learning techniques?}

Suppose we choose to evaluate $\mathcal E$ by forming and solving the adjoint problem again on a (globally) uniformly refined mesh and substituting its solution in place of the true adjoint solution.
This accurate -- but still approximate -- approach allows us to reliably produce a high quality representation of the error estimate.
However, it is also one of the most computationally expensive means of doing so (see the discussion in \cite[Section 7.3]{Wal21}).
It requires additional data structures, namely a mesh hierarchy and a means for transferring solution data between its levels.
The subscript $h$ in the notation $\mathcal H_h$ for the (`current') base mesh refers to the element sizing function, so let $\mathcal H_{h/2}$ denote the refined mesh generated by halving each element's size in each direction, e.g.~splitting a triangular element into four triangles of equal area in what is sometimes referred to as `red' refinement.
The associated function spaces are denoted $V_h$ and $V_{h/2}$ and satisfy $V_h\subset V_{h/2}$, for the Lagrange elements considered in this paper.
Further, denote by $u^*_{h/2}\in V_{h/2}$ the adjoint solution computed on $\mathcal H_{h/2}$, which we use to approximate $u^*$.
Approximating $\mathcal E_K$ amounts to the computation
\begin{equation}\label{eq:background:eval}
    \widetilde{\mathcal E_K}:=\rho\left.\left(P_{h/2}[u_h],u^*_{h/2}-P_{h/2}[u^*_h]\right)\right|_K,
    \qquad K\in\mathcal H_{h/2},
\end{equation}
where $P_{h/2}:V_h\rightarrow V_{h/2}$ transfers into the enriched space.
The weak residual is computed in this space.
The error indicator may then be approximated in the base space by applying a projection operator $\Pi_h:V_{h/2}\rightarrow V_h$:
\begin{equation}
    \widetilde{i_h}:=\Pi_h\left[\sum_{K\in\mathcal H_{h/2}}\widetilde i_K\right],\qquad
    \widetilde i_K:=\left\{\begin{array}{cc}
        \widetilde{\mathcal E_K} & \mathbf x\in K \\
        0 & \mathbf x\not\in K
    \end{array}\right..
\end{equation}
The reason that we transfer the error indicator information back down to the base space is that we are primarily motivated by mesh adaptation, which is driven by such information.

In this work, we choose $P_{h/2}$ to be a prolongation operator (which is lossless for the finite elements we consider) and $\Pi_h$ to be a conservative projection.
As such, successfully emulating the error estimate evaluation using machine learning is about more than just generating a better approximation of the adjoint solution; it also involves encoding information related to the enriched space and the associated interpolation operators.
However, at no point in the data-driven approach are these spaces actually constructed.

A na\"ive implementation of $u^*_{h/2}$ in (\ref{eq:background:eval}) involves solving the \emph{forward} problem in the enriched space and then forming the adjoint problem by linearising about its solution.
The additional cost of solving the enriched forward problem can be significant, especially for nonlinear problems.
In order to avoid this step during training, we calculate $u^*_{h/2}$ in an approximate sense: form the adjoint problem in the enriched space by linearising about \emph{the forward solution prolonged from the base space}, $P_{h/2}[u_h]$.
It was shown in \cite[Section 7.5]{Wal21} that this approach yields similarly effective error estimates to the na\"ive approach, at a dramatically reduced computational cost, since prolongation is orders of magnitude cheaper than global PDE solves.

\subsection{Tidal Turbine Modelling}\label{subsec:background:turbine}

In this work, we take as our physical demonstration scenario the depth-averaged modelling of coastal ocean flows using the \emph{shallow water equations}.
This hydrostatic model is derived from the Navier-Stokes equations and accounts for nonlinear advection, viscous forces, a free upper surface inducing pressure gradient like forces, and bed friction (or `drag') due to sea bed roughness.
In this setting, the spatial domain is two-dimensional: $\Omega\subset\mathbb R^2$.
The prognostic variables are horizontal velocity $\mathbf u:\Omega\rightarrow\mathbb R^2$ and free surface elevation $\eta:\Omega\rightarrow\mathbb R$.
We consider the steady-state shallow water equations in the non-conservative form,
\begin{equation}\label{eq:background:turbine:sw}
    \mathbf u\cdot\nabla\mathbf u
    +g\nabla\eta
    +C_{\mathcal D}\frac{\|\mathbf u\|\mathbf u}{\eta+b}
    =\nabla\cdot(\underline{\boldsymbol\nu}\nabla\mathbf u),\qquad
    \nabla\cdot((\eta+b)\mathbf u)
    =0,
\end{equation}
where $g=9.81\,\mathrm{m\,s}^{-2}$ is gravitational acceleration, $b$ [$\mathrm m$] is the bathymetry (sea floor topography), $C_{\mathcal D}$ is a dimensionless drag coefficient and $\underline{\boldsymbol\nu}$ [$\mathrm m^2\,\mathrm s^{-1}$] is the viscosity tensor.
For the purposes of this paper, we assume an isotropic viscosity coefficient, defined by a positive scalar field, $\underline{\boldsymbol\nu}=\nu\underline{\bf I}$.

Tidal stream turbines are comprised of blades and nacelle positioned on a vertically oriented shaft, where the plane of blade rotation is perpendicular to the horizontal component of the fluid flow.
As such, they are inherently 3D objects.
However, it is possible to represent them in a 2D setup using a drag parametrisation, with the presence of a turbine being modelled by a locally artificially increased drag coefficient.
As such, we decompose $C_{\mathcal D}=C_{\mathcal B}+C_{\mathcal T}$ into a constant background drag $C_{\mathcal B}=0.0025$ due to sea bed friction, and a spatially-varying drag due to the presence of turbines,
\begin{equation}
    C_{\mathcal T}=\sum_{T\in\mathcal T}C_T\mathbbm1_T.
\end{equation}
Here $\mathcal T$ denotes the set of all turbines.
In a 2D model, turbine $T\in\mathcal T$ is parametrised over its footprint, which is indicated by $\mathbbm1_T$ and has area $A_T^{\mathrm{footprint}}$.
Clearly, the turbines extract power based on the area $A_T^{\mathrm{swept}}$ that they sweep in the vertical, not their footprint.
As such, we apply the scaling
\begin{equation}
    C_T=\frac12\frac{A_T^{\mathrm{swept}}}{A_T^{\mathrm{footprint}}}c_T,
\end{equation}
where the so-called \emph{thrust coefficient} $c_T$ is related to details of the manufacture and operation of the turbines.
We assume a constant thrust coefficient of 0.8 herein, for simplicity.
An additional correction is applied to account for the fact that the thrust coefficient is defined in terms of the upstream velocity, rather than the depth-averaged value at the turbine location which is the value immediately available in the shallow water model (see \cite{KP16} for further details).

For the purposes of this paper, we take the QoI to be the power output of the tidal farm, a proxy of which may be computed as
\begin{equation}
    J(\mathbf u,\eta)=\int_\Omega\rho\,C_{\mathcal T}\|\mathbf u\|^3\;\mathrm dx,
\end{equation}
where $\rho=1030.0\,\mathrm{kg\,m}^{-3}$ is the assumed density of seawater.
The power output has units of Watts.

The shallow water equations (\ref{eq:background:turbine:sw}) are solved using the coastal ocean model \emph{Thetis} \cite{Thetis}.
We choose to discretise using the mixed discontinuous-continuous $\mathbb P1_{DG}-\mathbb P2$ finite element pair, first introduced in \cite{CH09}.
The corresponding weak formulation can be found in \cite[Section 3.1]{WB20}.
The resulting nonlinear systems are solved using a Newton method with line search, with the underlying linear systems solved by application of a full LU decomposition as a preconditioner \cite{MUMPS01,MUMPS02}.

Goal-oriented error estimates for the discretisation described above were derived in \cite[Section 3.2]{WB20}.
We refer to that work for details.

\section{Methodology}\label{sec:meth}

\subsection{Inputs and Outputs}\label{subsec:meth:io}

As discussed above, we seek to emulate DWR error estimation using a machine learning approach on an element-by-element basis.
This means that we seek to map inputs related to a particular mesh element $K$ of the base mesh to the associated error contribution, $\mathcal E_K$.
As such, for each element, there is a single real number to be computed.
There are a number of inputs that we deem to be important, as detailed in the following.
All of them are real-valued, meaning that we effectively seek to obtain a mapping
\begin{equation}\label{eq:meth:io:e2n}
    \mathrm{E2N}:P\rightarrow\mathbb R,\qquad P\subseteq\mathbb R^n,
\end{equation}
where $n$ is the number of scalar input parameters.

\paragraph{Coarse DWR approximation}

Our objective is to emulate the DWR error estimator (\ref{eq:background:eval}) without constructing enriched finite element spaces.
A na\"ive way to do this is to simply replace the true adjoint solution with the finite element adjoint solution on the base mesh\footnote{Whilst (\ref{eq:meth:io:estimator}) should sum to zero over all mesh elements, since (\ref{eq:background:weak_form}) holds with $v=u^*_h$, this does not mean that the individual elemental contributions are zero.}:
\begin{equation}\label{eq:meth:io:estimator}
    \rho(u_h,u^*_h)|_K.
\end{equation}
This is a low accuracy approximation of a DWR indicator, but is a potentially useful input feature for the neural network, which is readily calculable on the base mesh.

\paragraph{Physics-based parameters}

The next set of input parameters that we consider relate to the flow physics.
Given that the gravitational acceleration and seawater density are held fixed across test cases, the remaining parameters defining (\ref{eq:background:turbine:sw}) are the viscosity, drag coefficient and bathymetry.
In the case of spatially varying parameter fields, these are averaged over the element to provide scalar inputs to the element-wise network.

\paragraph{Mesh element parameters}

We also consider a number of input parameters related to the mesh element itself, as detailed in the following.
In this work we restrict attention to meshes comprised of triangular elements.

Mesh element $K$ may be obtained from the reference element $\widehat K$ using an affine transformation with Jacobian $\underline{\bf J}_K\;$\footnote{We assume $\widehat K$ to be the right-angled triangle formed by vertices $(0,0)$, $(0,1)$ and $(1,0)$. This is in agreement with the convention in Firedrake.}.
The matrix $\underline{\bf J}_K^T\underline{\bf J}_K$ is symmetric positive-definite (SPD), by construction.
Symmetry implies the orthogonal eigendecomposition,
\begin{equation}\label{eq:meth:io:decomp}
    \underline{\bf J}_K^T\underline{\bf J}_K
    =\underline{\bf V}_K\underline{\boldsymbol\Lambda}_K\underline{\bf V}_K^T.
\end{equation}
Positive-definiteness implies that the eigenvalues $\Lambda_K=\mathrm{diag}(\lambda_{1,K},\lambda_{2,K})$ are both positive, meaning we can define $h_1:=1/\sqrt{\lambda_{1,K}}$ and $h_2:=1/\sqrt{\lambda_{2,K}}$.
Assume w.l.o.g.~that the eigenvalues are in ascending order, so that $h_1\geq h_2>0$.

A convenient geometric interpretation for (\ref{eq:meth:io:decomp}) is in terms of the minimal ellipse containing the triangle's vertices.
Such an ellipse is shown in Figure \ref{eq:meth:io:decomp}, for an arbitrary triangle drawn in blue.
The eigenvectors represent the major semi-axes of the ellipse, with $h_1$ being the magnitude of the first and $h_2$ being the magnitude of the second.
We define the parameter $\theta\in[0,2\pi)$ to be the argument of the first eigenvector.
\begin{figure}
    \begin{floatrow}
    \ffigbox{
        \begin{tikzpicture}[scale=0.6]
            \draw [black!80, thick, ->] (0,0) -- (5,0) node[right=2pt] {$\widehat{\mathbf x}$};
    		\draw [Cyan, very thick, rotate=20] (3.5,-0.9682458365518543) -- (0,2);
    		\draw [Cyan, very thick, rotate=20] (0,2) -- (-3.5,-0.9682458365518543);
    		\draw [Cyan, very thick, rotate=20] (-3.5,-0.9682458365518543) -- (3.5,-0.9682458365518543);
    		\draw [gray, rotate=20] (4,0) arc (0:360:4cm and 2cm);
    		\draw [ForestGreen, thick, ->, rotate=20] (0,0) -- (4,0) node[right=2pt] {\color{ForestGreen}{$h_1=\frac1{\sqrt{\lambda_1}}$}};
    		\draw [ForestGreen, thick, ->, rotate=20] (0,0) -- (0,2) node[above=2pt] {\color{ForestGreen}{$h_2=\frac1{\sqrt{\lambda_2}}$}};
    		\draw [green, thick] (2.2,0) arc (0:20:4cm and 2cm);
    		\node [green, font=\small] at (1.5,0.25) {$\theta$};
    	\end{tikzpicture}
    }{
        \caption{
            Ellipse representation of the quantities $h_1$, $h_2$ and $\theta$ related to a triangular element.
            The vector $\widehat{\mathbf x}$ points in the direction of the positive $x$-axis.
        }
        \label{fig:meth:io:triangle}
    }
    \ffigbox{
        \begin{tikzpicture}[scale=2]
    		\draw [Cyan, very thick] (-0.5,0) -- (0.5,0) -- (0,0.866025) -- (-0.5,0);
    		\draw [Cyan, very thick] (1,0) -- (2,0) -- (1.5,0.866025) -- (1,0);
    		\draw [Cyan, very thick] (-0.5,1) -- (0.5,1) -- (0,1.866025) -- (-0.5,1);
    		\draw [Cyan, very thick] (1,1) -- (2,1) -- (1.5,1.866025) -- (1,1);
    		\filldraw [green] (0,1.288675) circle (1pt);
    		\node [ForestGreen, font=\small] at (-0.5,1.75) {$\mathbb P0$};
    		\filldraw [green] (1.2,1.1) circle (1pt);
    		\filldraw [green] (1.8,1.1) circle (1pt);
    		\filldraw [green] (1.5,1.6) circle (1pt);
    		\node [ForestGreen, font=\small] at (1,1.75) {$\mathbb P1_{DG}$};
    		\filldraw [green] (-0.5,0) circle (1pt);
    		\filldraw [green] (0.5,0) circle (1pt);
    		\filldraw [green] (0,0.866025) circle (1pt);
    		\node [ForestGreen, font=\small] at (-0.5,0.75) {$\mathbb P1$};
    		\filldraw [green] (1,0) circle (1pt);
    		\filldraw [green] (1.5,0) circle (1pt);
    		\filldraw [green] (2,0) circle (1pt);
    		\filldraw [green] (1.75,0.4330127) circle (1pt);
    		\filldraw [green] (1.5,0.866025) circle (1pt);
    		\filldraw [green] (1.25,0.4330127) circle (1pt);
    		\node [ForestGreen, font=\small] at (1,0.75) {$\mathbb P2$};
    	\end{tikzpicture}
    }{
        \caption{
            Diagrams representing the degrees of freedom of various finite elements.
            If a node is drawn on an edge then this implies that continuity is enforced with a neighbouring element.
        }
        \label{fig:meth:io:dofs}
    }
    \end{floatrow}
\end{figure}

With the above quantities, we can encapsulate the size of the element with $d:=h_1h_2$, the orientation of the element with $\theta$ and the shape of the element with the aspect ratio $s:=h_1/h_2$.
These three parameters uniquely define the element.
However, as discussed in \cite{FC21}, it is potentially problematic that orientations $\theta\approx0$ and orientations $\theta\approx2\pi$ are strongly correlated, even though this is not apparent in polar coordinates.
If this parameter were used as input to a neural network, it could hinder its performance.
Instead, we follow the recommendations of \cite{FC21} by replacing $\theta$ and $s$ with, respectively,
\begin{equation}\label{eq:meth:io:inputs}
    \frac1s\cos^2\theta+s\sin^2\theta
    \qquad\text{and}\qquad
    \left(s-\frac1s\right)\sin\theta\cos\theta.
\end{equation}

In addition to the above mesh-based parameters, we also include a parameter that integrates unity over the intersection of the element boundary and the domain boundary:
\begin{equation}\label{eq:meth:io:bnd}
    \int_{\partial K\cap\partial\Omega}\;\mathrm ds.
\end{equation}
Clearly, (\ref{eq:meth:io:bnd}) is zero if the element does not touch the domain boundary.
Otherwise, it has a positive value, whose magnitude is determined by the length of the shared boundary segment.

\paragraph{Solution fields}

Given that we seek to perform goal-oriented error estimation, there are many reasons the forward and adjoint solutions at each DoF make valuable choices of inputs.
We can assume that the adjoint solution from the base space provides a reasonable first approximation to the enriched version.
As such, it may well be key to guiding the network towards emulating the enrichment process.
Perhaps more importantly, the adjoint solution encodes the QoI, which is ultimately what we would like to accurately capture.
The approximate forward solution is potentially important because the adjoint solve in the enriched space involves a linearisation about its prolongation.

An average value of one of the solution fields over an element is typically insufficient information for the purposes of goal-oriented error estimation, since intra-element variation often plays an important role.
As such, the input data should reflect this variability.
One option is to simply provide the DoF values.
For the (relatively simple) finite elements used in this paper, all of the DoFs are given by point evaluations.
Figure \ref{fig:meth:io:dofs} represents the layouts of these DoFs on an equilateral triangle.
The nodal positions are geometrically accurate for the $\mathbb P1$ and $\mathbb P2$ diagrams.
For the $\mathbb P0$ case, the single DoF is associated with the cell.
The locations of the $\mathbb P1_{DG}$ DoFs are identical to the $\mathbb P1$ case (at the corners), it is just that inter-element continuity is not enforced.

A problem with simply supplying the DoFs, however, is that the ordering of the solution field DoFs on each element is essentially arbitrary in the finite element package used in this work.
This means that the link between the value of a DoF and its location is lost.
To combat this issue, we choose the input data for a field $f\in\mathbb Pp_{DG}$ (noting that $\mathbb Pp\subseteq\mathbb Pp_{DG}$) to be its value at the element centroid, $\mathbf x_K$ and the corresponding values of up to its $p^{th}$ derivatives.
For mixed derivatives, the values are averaged across orderings.
In particular, for a degree 2 field, we take the quantities
\begin{equation}
    f(\mathbf x_K),
    \:\frac{\partial f}{\partial x}(\mathbf x_K),
    \:\frac{\partial f}{\partial y}(\mathbf x_K),
    \:\frac{\partial^2f}{\partial x^2}(\mathbf x_K),
    \:\frac12\left(\frac{\partial^2f}{\partial x\partial y}(\mathbf x_K)+\frac{\partial^2f}{\partial y\partial x}(\mathbf x_K)\right),
    \:\frac{\partial^2f}{\partial x^2}(\mathbf x_K).
\end{equation}

Because of the averaging, a degree $p$ field gives rise to $\frac12(p+1)(p+2)$ input data points (the $p^{th}$ triangular number), which coincides with the DoF count on a triangular element.
For a $\mathbb P1_{DG}-\mathbb P2$ finite element pair, there are twelve DoFs per element (six velocity DoFs and six elevation DoFs).
These imply twelve input parameters for the forward solution and twelve for the adjoint.

\paragraph{Summary}

Overall, we have one input parameter that approximates the DWR estimator itself on the coarse mesh (see eq. (\ref{eq:meth:io:estimator})), three input parameters related to the ``physics'' of the problem being solved, four related to the mesh element and 24 input parameters related to the forward and adjoint solution fields.
This gives 32 input parameters in total.

\subsection{Training}\label{subsec:meth:train}

We consider a finite set of test cases, the majority of which are used for training the neural network (i.e.~\emph{supervised learning}) and the remainder being reserved as `unseen' problems.
For each test case, training data is generated by applying goal-oriented error \emph{estimation} (involving the application of the costly enrichment methods) on an initial mesh constructed using \emph{gmsh} \cite{gmsh}.
In addition, we apply goal-oriented mesh \emph{adaptation} based on the error indicators, until convergence of the QoI or mesh element count (see \cite[Algorithm 3]{Wal21}).
A minimum of three mesh adaptation iterations is used, which we have found to be sufficient for mesh anisotropy to become fully developed.
Training data is extracted from these first three adaptation iterations, since they are guaranteed to exist for all such test cases.

\subsection{Network Architecture}\label{subsec:meth:nn}

\begin{figure}
    \begin{floatrow}
    \capbtabbox{
        \begin{tabular}{c|c}
            Network type                & Fully connected\\
            Number of inputs            & 32\\
            Number of hidden layers     & 1\\
            Number of hidden neurons    & 64\\
            Number of outputs           & 1\\
            Cost function               & Mean square error\\
            Activation function         & Sigmoid\\
            Preprocessing function      & Arctangent\\
            Learning rate               & 0.001 \\
            Number of epochs            & 2,000\\
            Batch size                  & 500\\
            Training/validation ratio   & 7:3\\
            Optimisation routine        & Adam\\
            Software                    & Pytorch\\
        \end{tabular}
    }{
        \caption{
            A table describing the neural network configuration and parameter values.
        }
        \label{tab:meth:nn:params}
    }
    \ffigbox{
        \includegraphics[width=0.5\textwidth, angle=90, origin=c, trim={470 120 80 200}, clip]{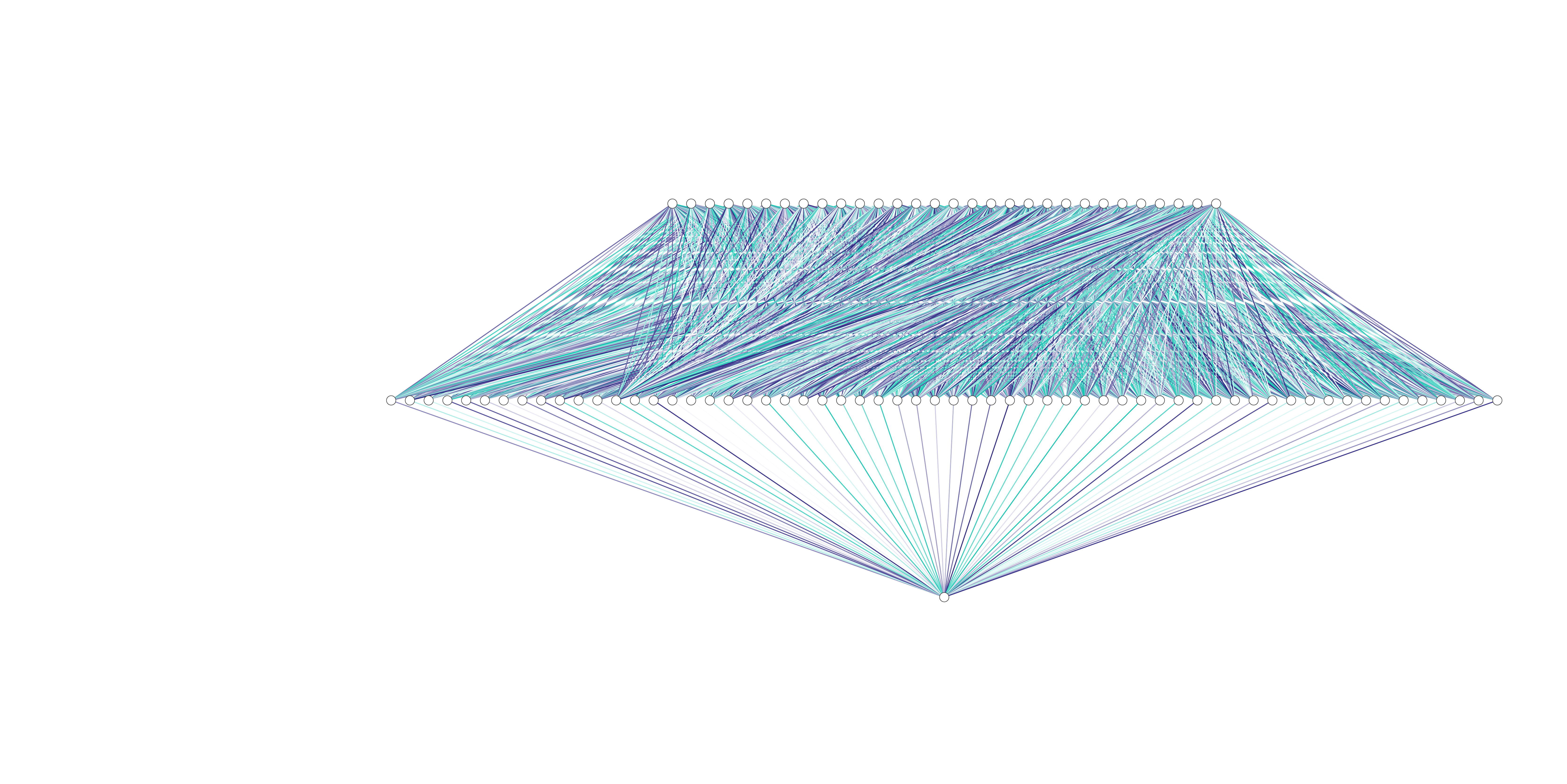}
    }{
        \caption{
            Graph representation of the network, obtained using \cite{Len19}.
        }
        \label{fig:meth:nn:net}
    }
    \end{floatrow}
\end{figure}

Table \ref{tab:meth:nn:params} details the parameter values used for the neural network.
We choose a feed-forward fully connected neural network with a single hidden layer, which is represented as a graph diagram in Figure \ref{fig:meth:nn:net}.
A standard sigmoid activation function is used.
All feature data is pre-processed using the arctangent function, which acts as a normalisation step and avoids biasing towards extreme values.
We opt to use a relatively simple network architecture for this preliminary work, with the aim of proving that it is possible to successfully emulate goal-oriented error estimation strategies without expending significant effort on optimisation.

The neural network and its training are implemented in Python using \emph{PyTorch} \cite{Pytorch}.
\emph{Adam} \cite{KB14} is used as the stochastic optimisation routine.

\subsection{Metric-based Mesh Adaptation}\label{subsec:meth:metric}

Whilst the network that we seek to construct relates to error estimation rather than mesh adaptation, we do use different input meshes to generate the training data, some of which are obtained following mesh adaptation.
Mesh adaptation is also used in Section \ref{sec:numexp}.
As such, we outline the approach in the following.

\paragraph{Background}

The metric-based framework was first introduced in \cite{GH91} and is a mesh adaptation approach based on \emph{Riemannian metric fields}.
A Riemannian metric, $\mathcal M=\{\underline{\bf M}(\mathbf x)\}_{\mathbf x\in\Omega}$, is a tensor field over the spatial domain $\Omega$, which takes the value of an SPD matrix at each point.
The metric provides a concept of mesh optimality which can in turn be used to drive a mesh adaptation process.
That is, at each step where a mesh modification is to occur, we query which of the available local operations yields the greatest increase in quality, as measured by a quality functional related to the metric.
(See \cite[Equation (5.10)]{Wal21} for example.)
For the 2D metric-based mesh adaptation (or mesh optimisation) tool used in this work, \emph{Mmg2d} \cite{Mmg}, the available operations are node insertion, node deletion, edge swapping and node movement.
A great advantage of the metric-based approach is that it enables control of the shape and orientation of elements, as well as their size, leading to the ability to generate highly anisotropic elements.
Riemannian metrics can also be viewed as continuous analogues for meshes (see \cite{LA11a,LA11b} for details) -- a duality with many useful insights for metric-based mesh adaptation.
We refer to the above citations and \cite{PU01} for further details on the metric-based framework.

\paragraph{Goal-Oriented Metrics}

It is the metric which drives the mesh adaptation process, so it must be carefully selected.
In this work, we use an anisotropic formulation based on the goal-oriented error indicators discussed in Subsection \ref{subsec:background:ee}.
This is taken from \cite{CP13} and is built upon the scaling,
\begin{equation}\label{eq:meth:metric:scaling}
    m(K):=\frac{|K|}{|\widetilde K|}
    \frac{\mathcal C\:\mathcal E_K^{\frac1{\alpha+1}}}{\sum_{K\in\mathcal H_h}\mathcal E_K^{\frac1{\alpha+1}}},
\end{equation}
where $|K|$ is the area of element $K$ and $\alpha\geq1$ is a parameter controlling the extent to which the adapted mesh captures multiple scales.
The parameter value $\alpha=1$ was found to be effective in \cite{WB20} and so is used here.
Metric complexity is the continuous analogue of the (discrete) mesh vertex count \cite{LA11a}, so the target metric complexity, which we denote by $\mathcal C$, effectively controls the number of vertices in the adapted mesh.
It was proved in \cite{Bra98} that the scaling (\ref{eq:meth:metric:scaling}) solves the optimisation problem of minimising interpolation error, subject to attaining the target metric complexity.

An element-wise anisotropic metric can be obtained by applying the scaling (\ref{eq:meth:metric:scaling}) to a matrix that encodes the element's shape and orientation.
In this work, we choose the matrix to be a combination of Hessians.
Each Hessian is recovered from a scalar component of the forward solution using Cl\'ement interpolation \cite{Cle75} (see \ref{sec:recovery} for details).
The combination is formed by taking the entry-wise mean.
Each Hessian is already symmetric, meaning their mean is, too.
Therefore, like (\ref{eq:meth:io:decomp}), it admits an orthogonal decomposition,
\begin{equation}\label{eq:meth:metric:decomp}
    \underline{\bf H}_K
    =\underline{\bf V}_K\underline{\boldsymbol\Lambda}_K\underline{\bf V}_K^T.
\end{equation}
Assuming w.l.o.g.~that the eigenvalues $\underline{\boldsymbol\Lambda}_K=\mathrm{diag}(\lambda_1,\lambda_2)$ are ordered such that they are ascending in magnitude, we consider a stretching factor, $s_K:=\sqrt{|\lambda_2/\lambda_1|}$ (much like the aspect ratio in Subsection \ref{subsec:meth:io}).
The anisotropic metric is formed as
\begin{equation}\label{eq:meth:metric:anisotropic}
    \underline{\bf M}_K^{\mathrm{anisotropic}}
    :=m(K)\:
    \underline{\bf V}_K\:
    \mathrm{diag}\left(s_K,\frac1{s_K}\right)\:
    \underline{\bf V}_K^T.
\end{equation}
Similarly as in Subsection \ref{subsec:meth:io}, (\ref{eq:meth:metric:anisotropic}) neatly decomposes the metric into components affecting size (the scaling $m(K)$), orientation (the eigenvector matrix $\underline{\bf V}_K$) and shape (the stretching factor $s_K$).

\paragraph{Steady-State Mesh Adaptation Algorithm}

\begin{figure}
    \centering
    \begin{tikzpicture}[scale=0.45, node distance = 1.5cm, auto]
        \node [block, font=\footnotesize] (fwd) {Solve forward on $\mathcal H_h^i$; evaluate QoI};
        \node [cloud, left of=fwd, font=\footnotesize] (input) {$\mathcal H_h^0$, $i:=0$};
        \node [decision, below of=fwd, font=\footnotesize] (check1) {QoI converged?};
        \node [block, below of=check1, node distance=3cm, font=\footnotesize] (adj) {Solve adjoint on $\mathcal H_h^i$};
        \node [cloud, right of=check1, node distance=3cm, font=\footnotesize] (stop) {Terminate};
        \node [special block, right of=adj, node distance=3cm, font=\footnotesize] (ee) {Evaluate error indicators on $\mathcal H_h^i$};
        \node [block, right of=fwd, node distance=3cm, font=\footnotesize] (inc) {Increment $i:=i+1$};
        \node [decision, right of=inc, font=\footnotesize] (check2) {Mesh converged?};
        \node [block, right of=ee, node distance=3cm, font=\footnotesize] (met) {Construct metric of complexity $\mathcal C^i$ on $\mathcal H_h^i$};
        \node [block, above of=met, node distance=3cm, font=\footnotesize] (adapt) {Adapt mesh as $\mathcal H_h^{i+1}$};
        \path [line,dashed] (input) -- (fwd);
        \path [line] (fwd) -- (check1);
        \path [line] (check1) -- node {\small yes}(stop);
        \path [line] (check1) -- node {\small no}(adj);
        \path [line] (adj) -- (ee);
        \path [line] (ee) -- (met);
        \path [line] (met) -- (adapt);
        \path [line] (adapt) -- (check2);
        \path [line] (check2) -- node {\small yes}(stop);
        \path [line] (check2) -- node {\small no}(inc);
        \path [line] (inc) -- (fwd);
    \end{tikzpicture}
    \caption{
        Diagram illustrating the fixed point iteration loop used for the mesh adaptation process applied to steady-state PDE problems.
        The error indication step is highlighted in a brighter blue than the other computation steps because it is the one that we seek to accelerate.
    }
    \label{fig:meth:metric:loop}
\end{figure}
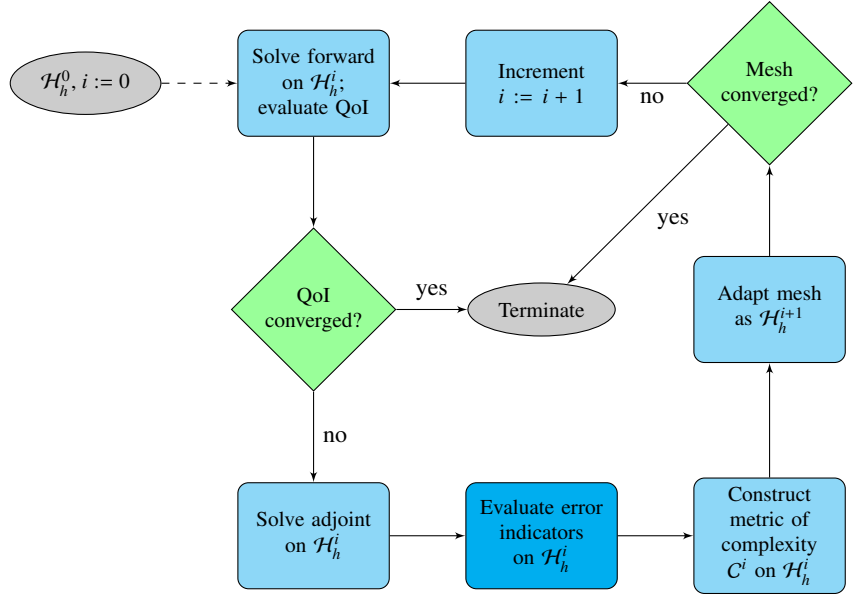

The anisotropic metric construction strategy is described above.
Building upon this, we outline the mesh adaptation algorithm in the following.
Figure \ref{fig:meth:metric:loop} provides a workflow representation.

We begin with an initial mesh, $\mathcal H_h^0$, where the superscript indicates the fixed point iteration.
Then, in iteration $i\in\mathbb N$, we begin by solving the forward problem in the base finite element space on $\mathcal H_h^i$ and thereby evaluate the QoI.
If it has not converged since the previous iteration, we progress to solve the adjoint equation in the base space.
The next step is to indicate errors, which is the only place where the data-driven approach differs from the standard goal-oriented approach.
Given error indicators, we then construct an anisotropic metric.
For steady-state problems, it is often beneficial to `spin up' the target metric complexity in this step, especially from the perspective of computational efficiency.
As such, $\mathcal C^i$ denotes the target complexity at step $i$, and is typically prescribed to plateau at the desired value after three iterations
(see \cite{WB21} for details).
Finally, we adapt (or optimise) the mesh based on the metric and check for convergence of its element count.
If the overall change in element count is sufficiently small then the algorithm is terminated.
Typically, the process is also terminated (with failure) at this point if a maximum iteration count is reached.
Otherwise, the process continues.
We find that three or more iterations are typically required in order for anisotropic mesh features to manifest themselves properly, so opt to turn off the convergence checks for the first three iterations, effectively imposing a `minimum iteration count'.

\section{Numerical Experiments}\label{sec:numexp}

\subsection{Problem Setup}\label{subsec:numexp:setup}

Goal-oriented error estimation and mesh adaptation were previously applied to steady-state shallow water modelling in \cite{WB20}.
That work focused on two test cases modelling flow around two turbines positioned in a compass-aligned channel with uniform inflow velocity from west to east.
In one case, the second turbine was positioned directly behind the first to maximise the effects of wake interaction.
In the other case, the turbines were offset to the north and south, so that the second turbine was not fully in the wake of the first.
This offsetting gave rise to a heightened power output from the turbines, in agreement with the literature \cite{DN14}.
By applying isotropic goal-oriented mesh adaptation, it was found that the power output could be more accurately captured than with (global) uniform refinement of the initial mesh.
Yet further improvements were found using anisotropic goal-oriented mesh adaptation.

The aim of the numerical experiments in this section is to replicate the improved convergence properties of the goal-oriented methods exhibited in \cite{WB20} using a data-driven variant, except at significantly reduced computational cost.
In order to achieve this, it is necessary to supply ample training data for the neural network.
Due to the design decision to work on an element-by-element basis, a significant amount of data is available in a single test case.
The amount of test data is increased by considering a variety of different test cases.
The hope is that this will also improve the generalisability of the network.
Yet more data is obtained by collecting features on adapted meshes, as well as the initial mesh.
For each test case, we consider the initial mesh and two adapted meshes, implying a total of three runs per test case.

The domain used in each setup is the same: the rectangle $\Omega=[1200\,\mathrm{m},500\,\mathrm{m}]$.
This is sufficient for the purposes of modelling steady-state flow in other domains, provided that their geometries do not introduce flow physics that the network did not ``see'' during training.
The reason for this is the fact that we work on an element-by-element basis, meaning that global information (such as domain geometry) is not available to the network.
However, in practice, almost any alteration of the domain geometry will give rise to different flow physics from those present in a straight channel and so we are limited to rectangular domains for now.
See Section \ref{sec:conc} for further discussion on this.

\begin{figure}[t]
    \centering
    \includegraphics[width=\textwidth, trim={10 0 10 10}, clip]{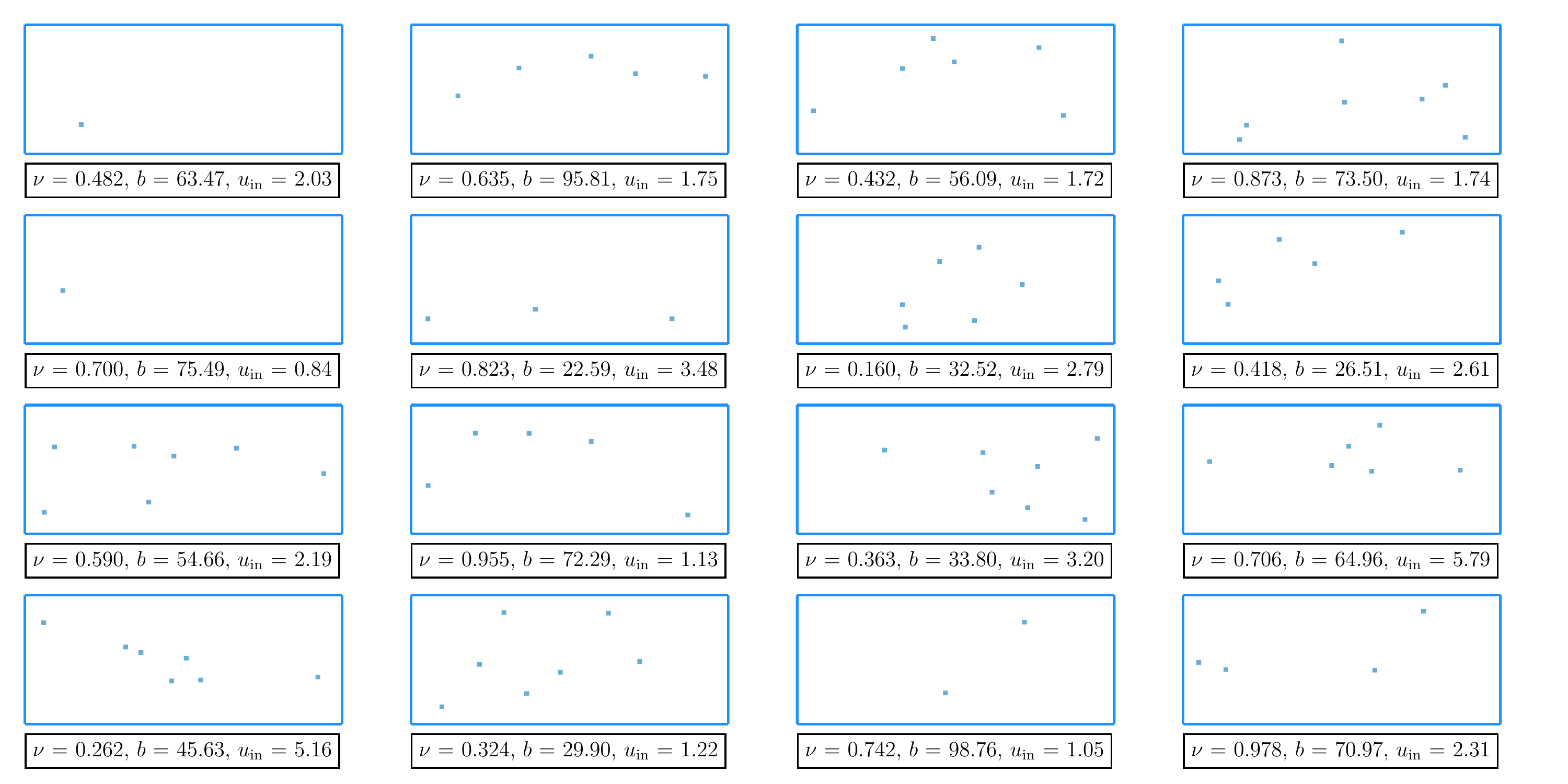}
    \caption{
        The first 16 setups used for training and validation (i.e.~`seen' scenarios).
        Each plot shows the associated domain boundary and turbine footprints, along with annotated viscosity coefficient, bathymetry and inflow velocity.
    }
    \label{fig:numexp:setup:train}
\end{figure}

Whilst the domains used for training are identical, the configuration of turbines within them are chosen randomly.
This reduces the amount of developer effort required to set up the training data and removes any human bias that might be introduced by choosing particular configurations.
It also means that the training data can be easily scaled up to an arbitrary number of tidal farm configurations.
For each test case, a random number of turbines between one and eight (inclusive) is sampled from an integer uniform distribution, i.e.~$U\{1,8\}$.
For each such turbine, $x$- and $y$-coordinates are sampled from the continuous uniform distributions $U(50,1150)$ and $U(50,450)$, respectively (i.e.~at least $50\,\mathrm m$ from any of the domain boundaries in the $\ell_\infty$ norm).
A proposed turbine location is accepted if it is at least $50\,\mathrm m$ from any other turbines and rejected otherwise.
The selection procedure iterates until all turbines have been positioned in the farm.
Figure \ref{fig:numexp:setup:train} shows the first 16 randomly generated setups, out of 100 in total.
Given that we consider the initial mesh and an additional two adapted meshes per test case, feature data is extracted from a total of 300 meshes.
This gives rise to 711,307 elements across all of our experiments.
The feature and output data associated with these elements are divided into subsets for training and validation in the ratio 7:3.
\begin{figure}[t]
    \centering
    \includegraphics[width=0.48\textwidth, trim={10 0 10 10}, clip]{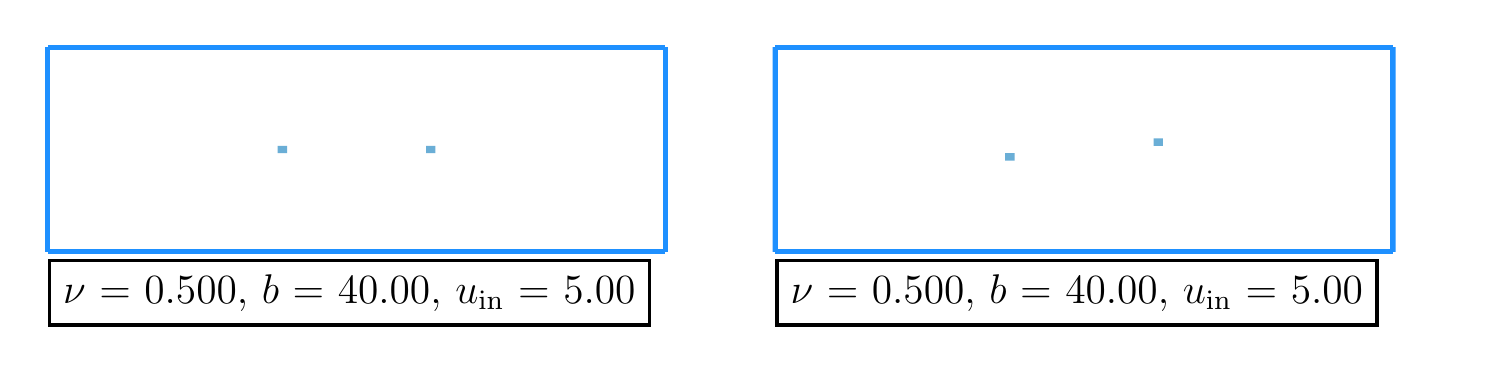}
    \caption{
        The two `aligned' and `offset' setups used for testing (i.e.~`unseen' scenarios).
        Each plot shows the associated domain boundary and turbine footprints, along with annotated viscosity coefficient, bathymetry and inflow velocity.
    }
    \label{fig:numexp:setup:test}
\end{figure}

It should be noted that positioning many turbines in a domain of constrained size will typically mean that goal-oriented mesh adaptation is less effective than it would be in the presence of fewer turbines or in the case of modelling a larger geographical region.
This is because there will likely be a higher proportion of the domain where high mesh resolution is required for an accurate QoI evaluation.
Consequently, the improvement over uniform refinement may be lessened for the same DoF count.
However, these factors do not mean that training the network on such scenarios is not of value.
In fact, the local approach means that more turbine interaction dynamics can be learnt than if only small turbine farms were considered.

In order to improve the generalisability of the network, we sample the viscosity, bathymetry and inflow speed parameters from uniform distributions, where each parameter is taken to be constant in space.
By doing so, we aim to increase the span of the explored portion of parameter space.
Whilst viscosity is a physical property of fluids, it also provides a parameter which can be artificially increased in order to stabilise a numerical model.
As such, the viscosity coefficient value is kept as $\mathcal O(10^{-1})$ across the test cases because this helps to maintain stability.
In detail, $b\sim U(20,100)$, $u_{\mathrm{in}}\sim U(0.5,6)$ and $\nu\sim U(0.1,1)$.
The values are annotated in Figure \ref{fig:numexp:setup:train}.
Another means of generalisation could be to choose spatially varying viscosity, bathymetry and inflow parameters.
Note that, whilst inflow speed is varied across the training data, it is not included as an input parameter for the network.
This is because we would like the network to be able to generalise to real scenarios where the inflow is so far away that its value is not of immediate importance, or where different inflow conditions are used.

\begin{figure}[t]
    \centering
    \begin{subfigure}{0.49\textwidth}
        \centering
        \includegraphics[width=\textwidth, trim={20 20 20 20}, clip]{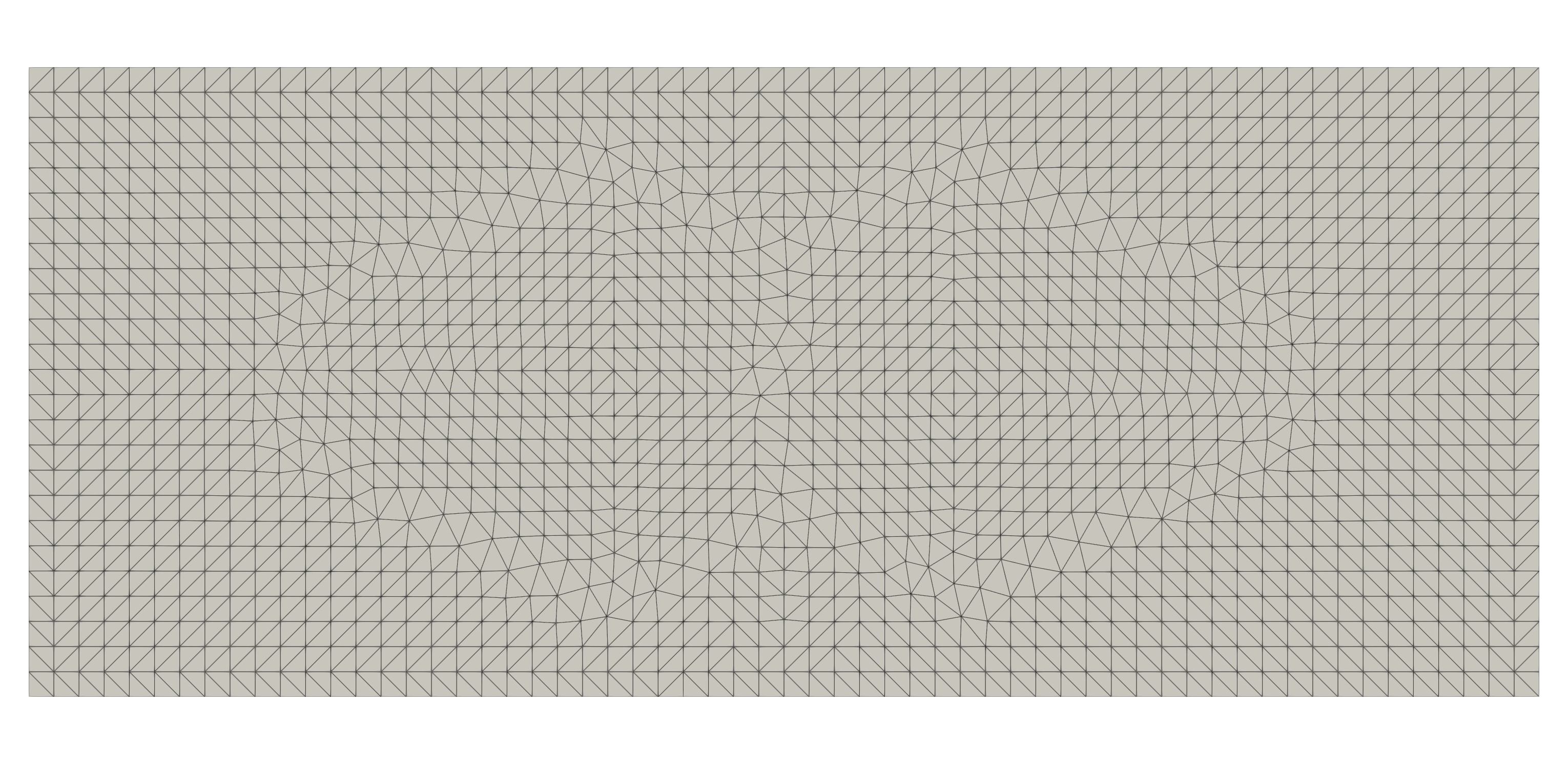}
        \caption{Aligned case}
        \label{subfig:numexp:setup:meshes:aligned}
    \end{subfigure}
    \begin{subfigure}{0.49\textwidth}
        \centering
        \includegraphics[width=\textwidth, trim={20 20 20 20}, clip]{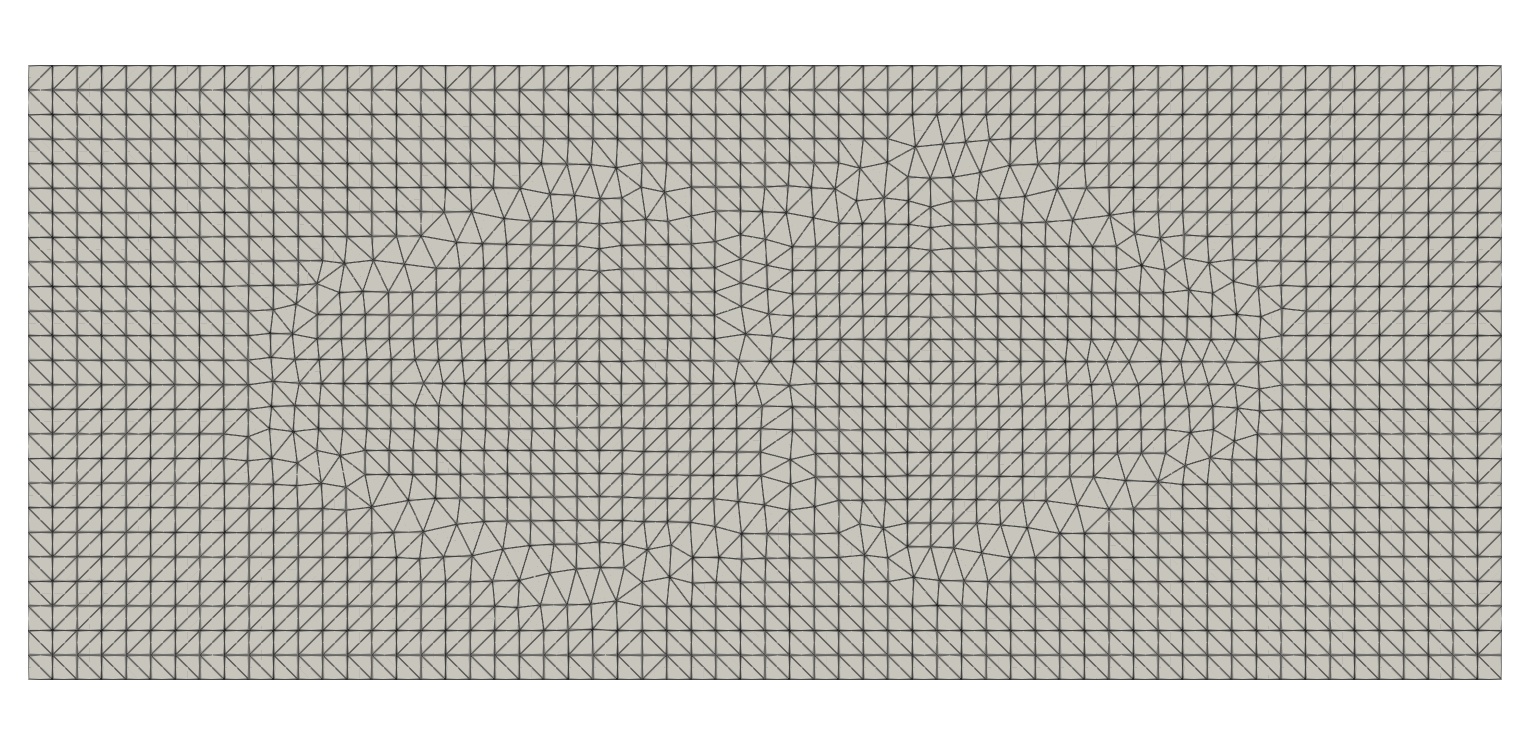}
        \caption{Offset case}
        \label{subfig:numexp:setup:meshes:offset}
    \end{subfigure}
    \caption{
        Quasi-uniform initial meshes used for the `aligned' and `offset' used for testing (i.e.~`unseen' scenarios).
    }
    \label{fig:numexp:setup:meshes}
\end{figure}

The test set is comprised of the two aligned and offset configurations from \cite{WB20}, which we seek to improve upon.
They are shown in Figure \ref{fig:numexp:setup:test}, along with the corresponding physical parameters.
Using these domain geometries, we generate quasi-uniform initial meshes using the `pack' algorithm within \emph{gmsh}, as shown in Figure \ref{fig:numexp:setup:meshes}.
In particular, we generate meshes that are as close as possible to a uniform mesh of right-angled triangles with edge length equal to the turbine diameter, under the constraint that the turbine footprints are to be explicitly meshed.
On these initial meshes, each turbine is meshed using two right-angled triangles.
The same quasi-uniform meshing strategy is used to generate initial meshes for the 100 training cases.

\subsection{Testing}\label{subsec:numexp:testing}

\paragraph{Error Indicators}

Before making a formal assessment of the data-driven method's performance, we first consider the error indicator fields that it generates, in order to confirm that they have similar structures to those generated by goal-oriented mesh adaptation.
\begin{figure}[t]
    \centering
    \includegraphics[width=0.6\textwidth, trim={300 700 370 500}, clip]{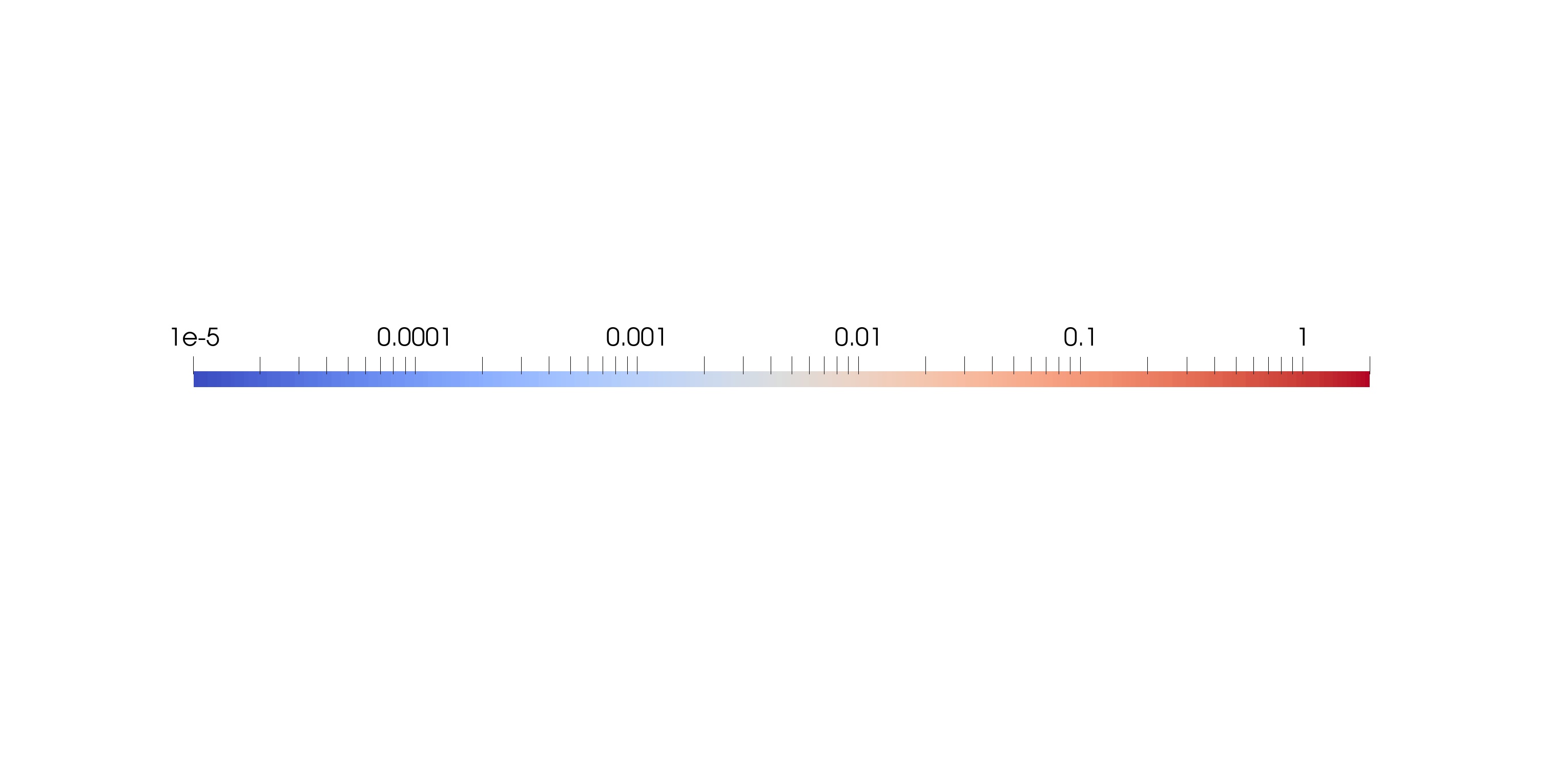}
    \begin{subfigure}{0.49\textwidth}
        \centering
        \includegraphics[width=\textwidth, trim={20 20 20 20}, clip]{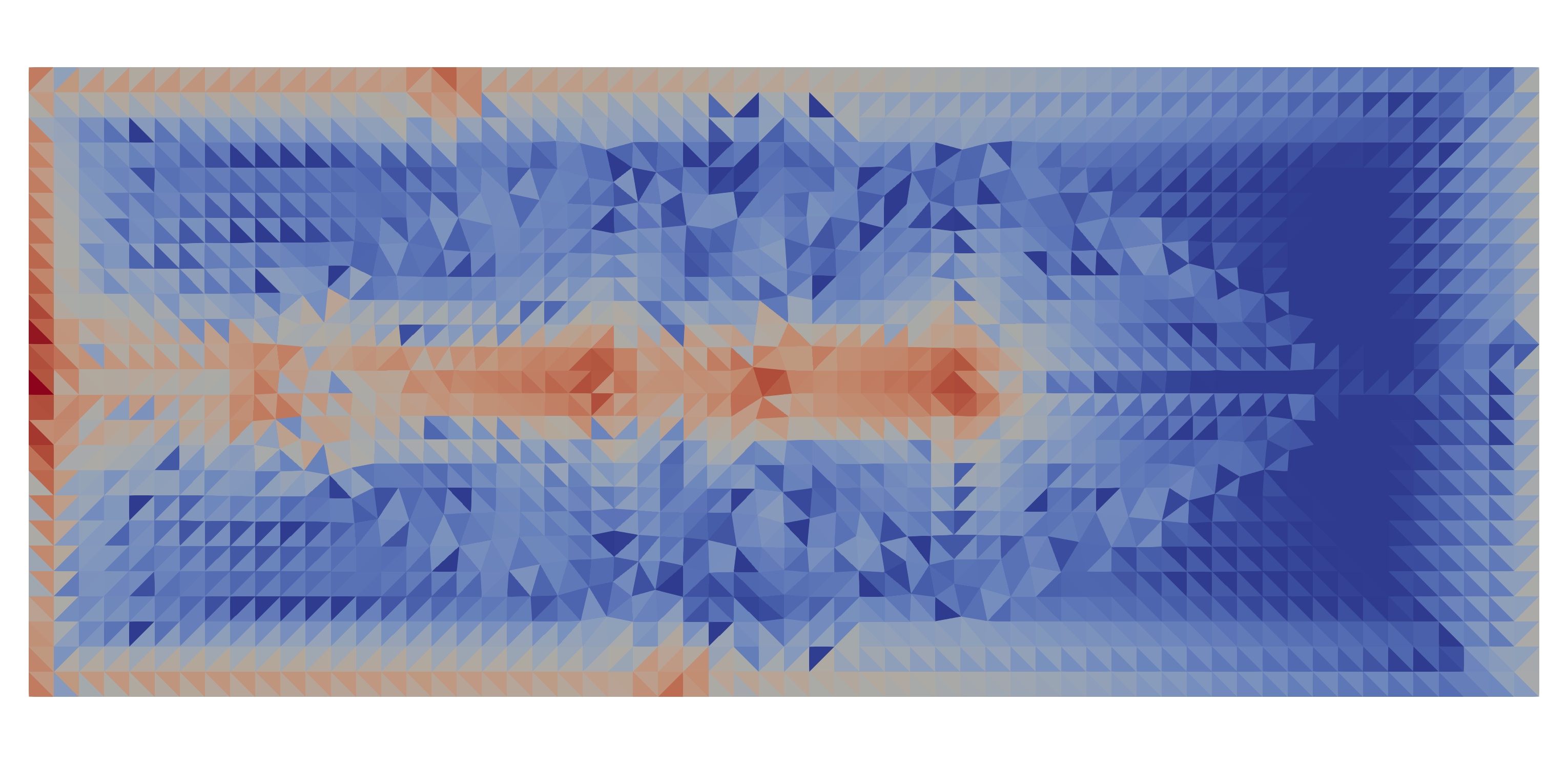}
        \caption{Aligned, goal-oriented}
        \label{subfig:numexp:testing:indicator:aligned:GO}
    \end{subfigure}
    \begin{subfigure}{0.49\textwidth}
        \centering
        \includegraphics[width=\textwidth, trim={20 20 20 20}, clip]{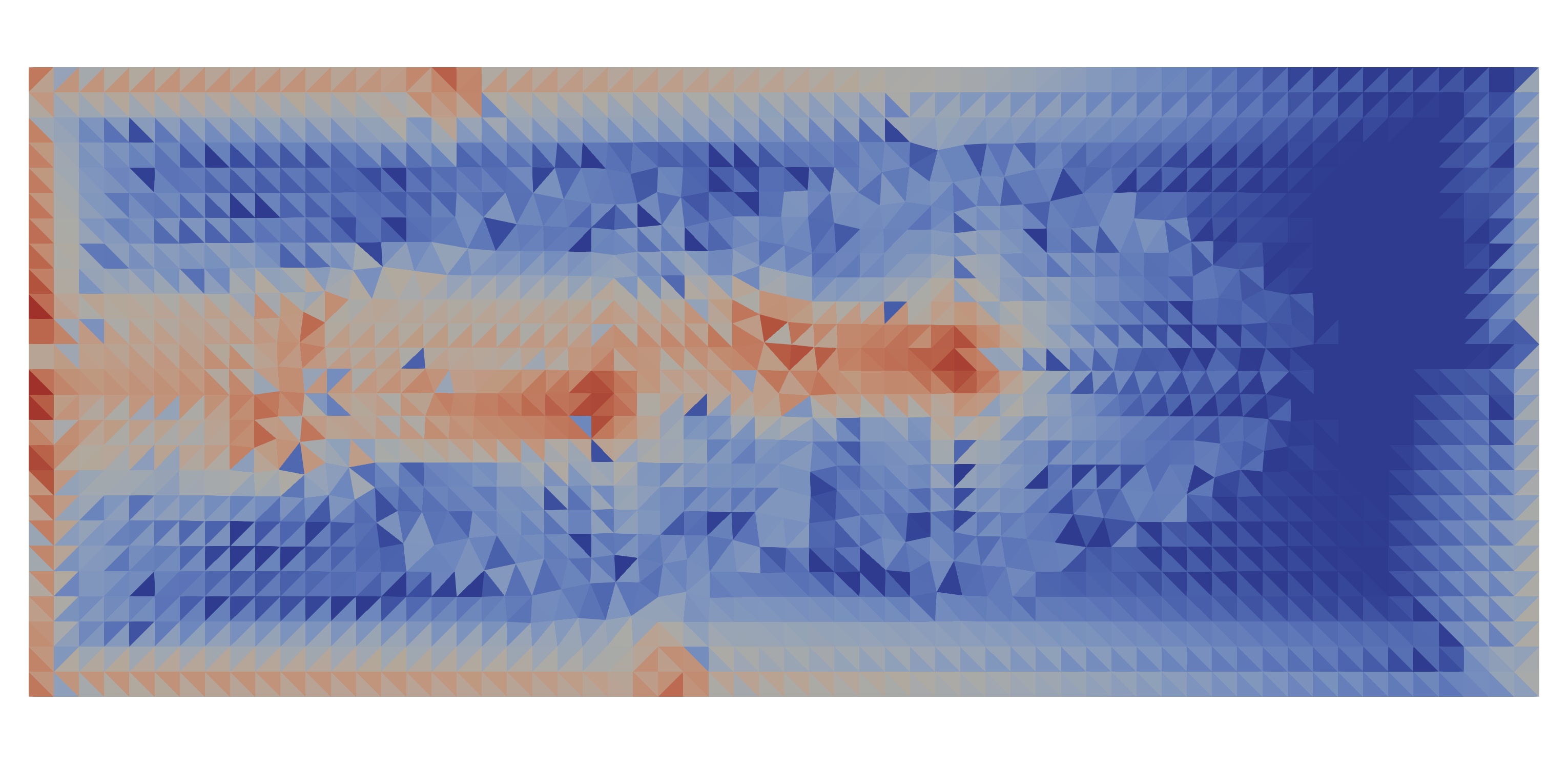}
        \caption{Offset, goal-oriented}
        \label{subfig:numexp:testing:indicator:offset:GO}
    \end{subfigure}
    \includegraphics[width=0.6\textwidth, trim={300 700 370 500}, clip]{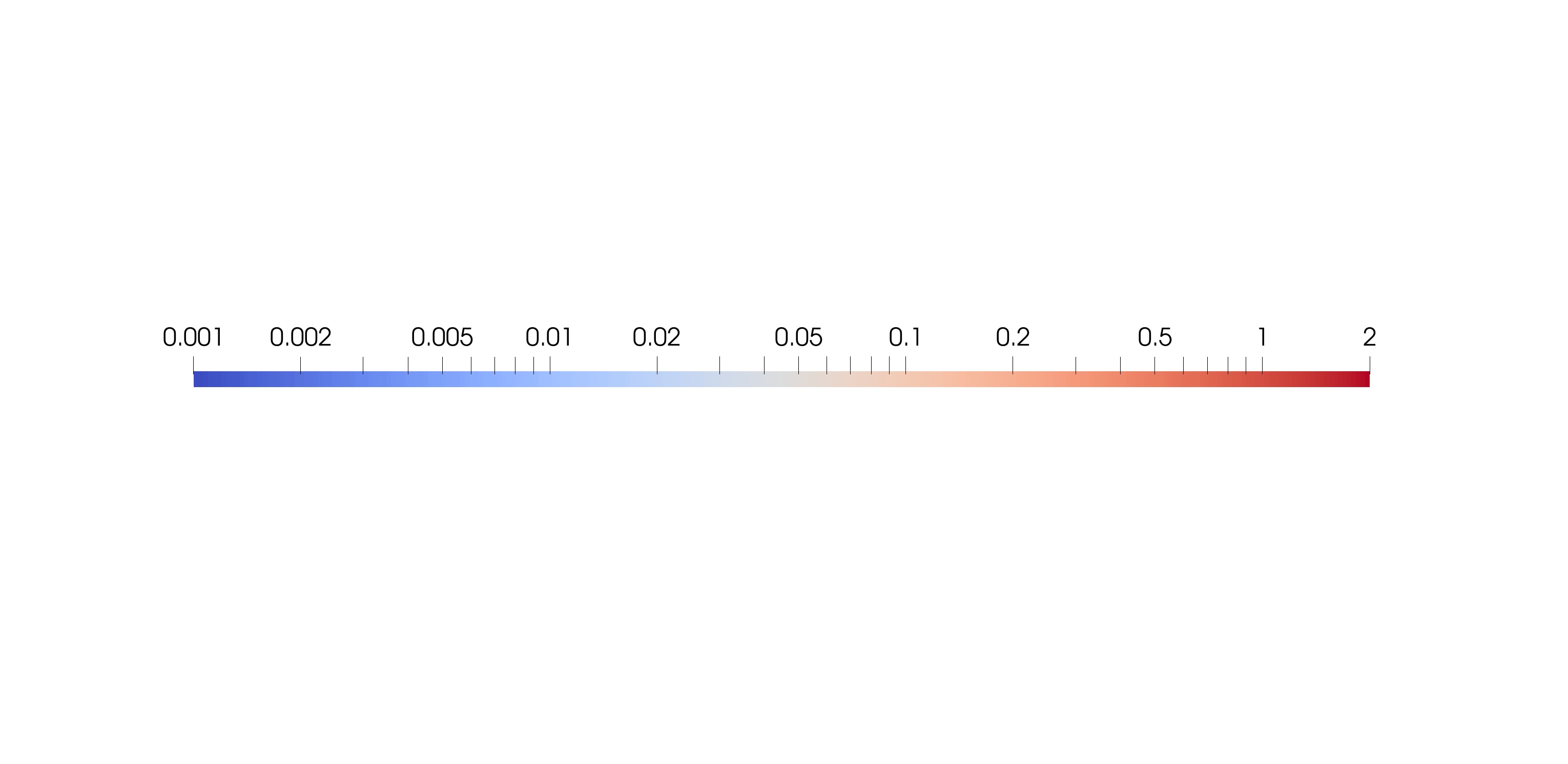}
    \begin{subfigure}{0.49\textwidth}
        \centering
        \includegraphics[width=\textwidth, trim={20 20 20 20}, clip]{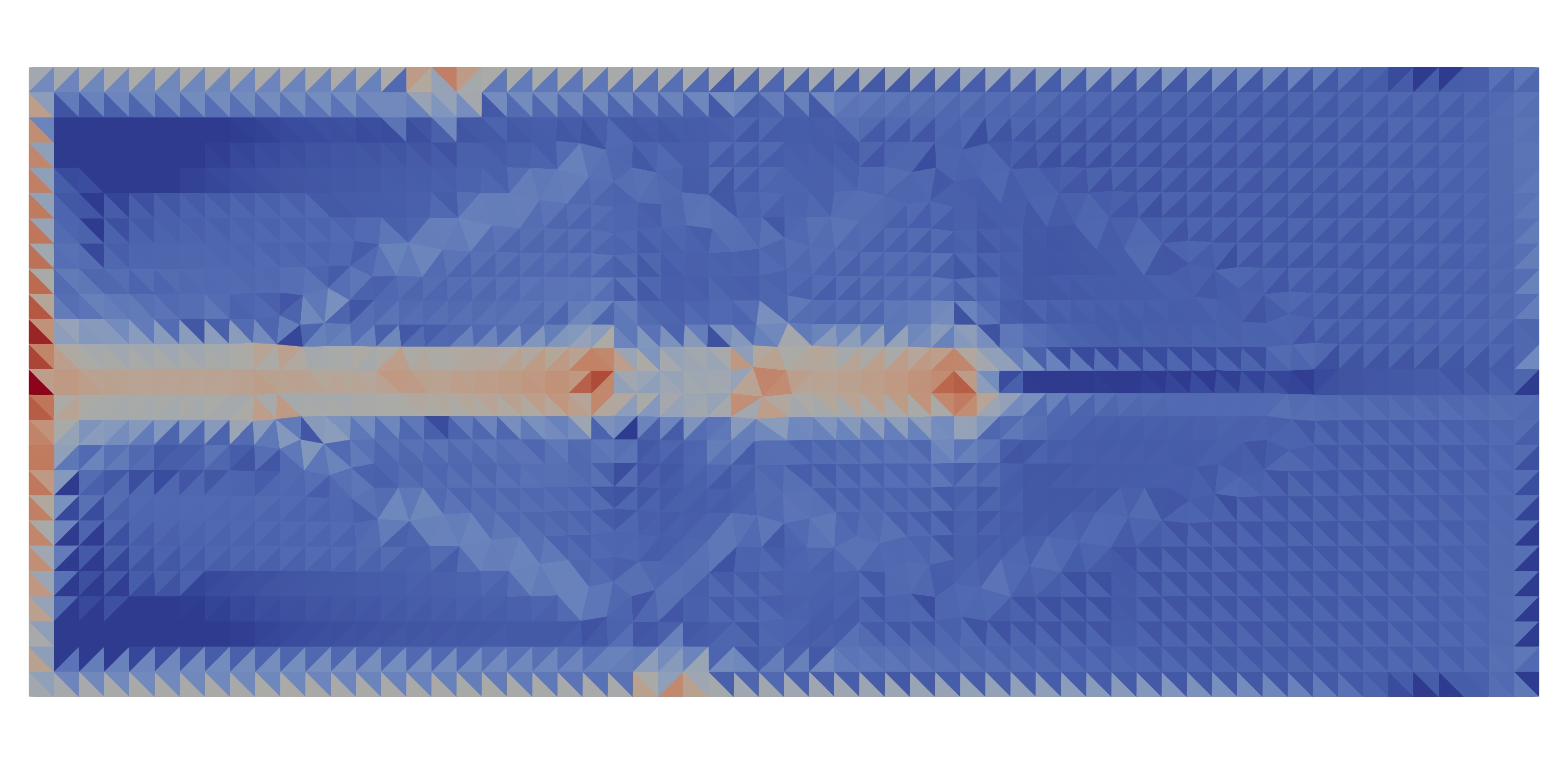}
        \caption{Aligned, data-driven}
        \label{subfig:numexp:testing:indicator:aligned:ML}
    \end{subfigure}
    \begin{subfigure}{0.49\textwidth}
        \centering
        \includegraphics[width=\textwidth, trim={20 20 20 20}, clip]{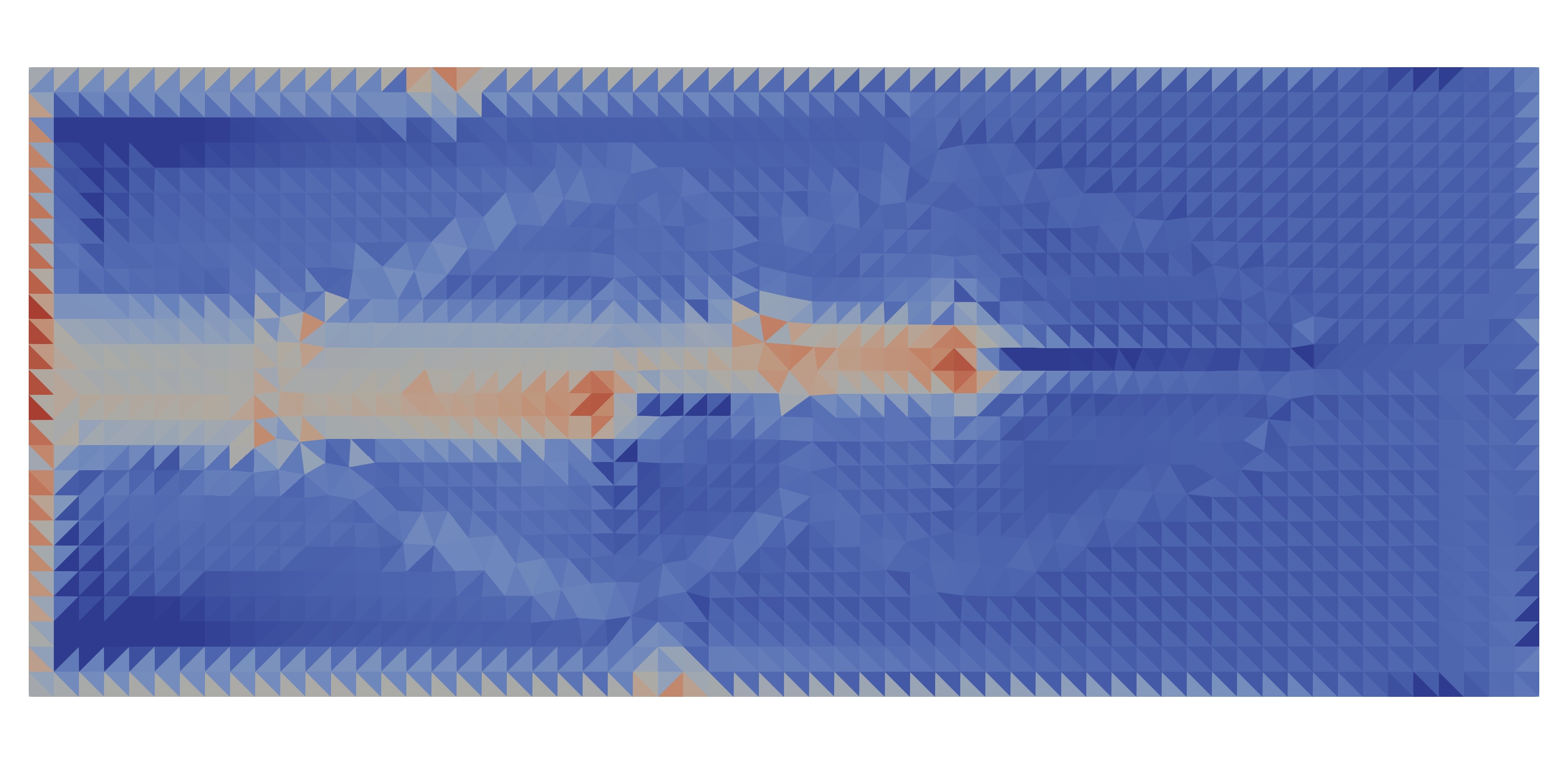}
        \caption{Offset, data-driven}
        \label{subfig:numexp:testing:indicator:offset:ML}
    \end{subfigure}
    \caption{
        Error indicator fields as computed on the `aligned' and `offset' initial meshes shown in Figure \ref{fig:numexp:setup:meshes}.
        Error indicators computed using standard goal-oriented mesh adaptation are shown at the top, whilst the corresponding data-driven indicators are shown at the bottom.
    }
    \label{fig:numexp:testing:indicator}
\end{figure}

The top row of plots in Figure \ref{fig:numexp:testing:indicator} show the error indicator fields according to the standard goal-oriented approach.
Contributions to the QoI error are estimated to be largest in magnitude at and upstream of the turbines.
This makes sense because the hydrodynamics in these regions significantly impact upon the QoI (power output) value.
Errors are also indicated along the domain boundary in the left-hand part of the domain.
These occur because the flow model, Thetis, does not impose boundary conditions exactly; it weakly enforces them, as is common practice in DG codes.
Error contributions are estimated to be smaller outside of the regions mentioned above, especially downstream of the turbine farm.
This makes sense because we have an advection-dominated problem, so the downstream hydrodynamics do not impact upon the QoI.

Consider now the bottom row of plots, which are generated using the data-driven approach.
Many of the important features appear to be captured well: the highest error contributions to the QoI are at and upstream of the turbines, errors are indicated along the boundary in the left-hand part of the domain and the lowest errors are indicated downstream of the turbine farm.
The indicator fields are not identical, however.
The most notable difference is that the magnitudes are often not that well captured, especially when the error estimate values are low.
In addition, the error indicator fields tend to be `smoother' than predicted by the standard goal-oriented approach.

\paragraph{Adapted Meshes}

\begin{figure}[t]
    \centering
    \begin{subfigure}{0.49\textwidth}
        \centering
        \includegraphics[width=\textwidth, trim={20 20 20 20}, clip]{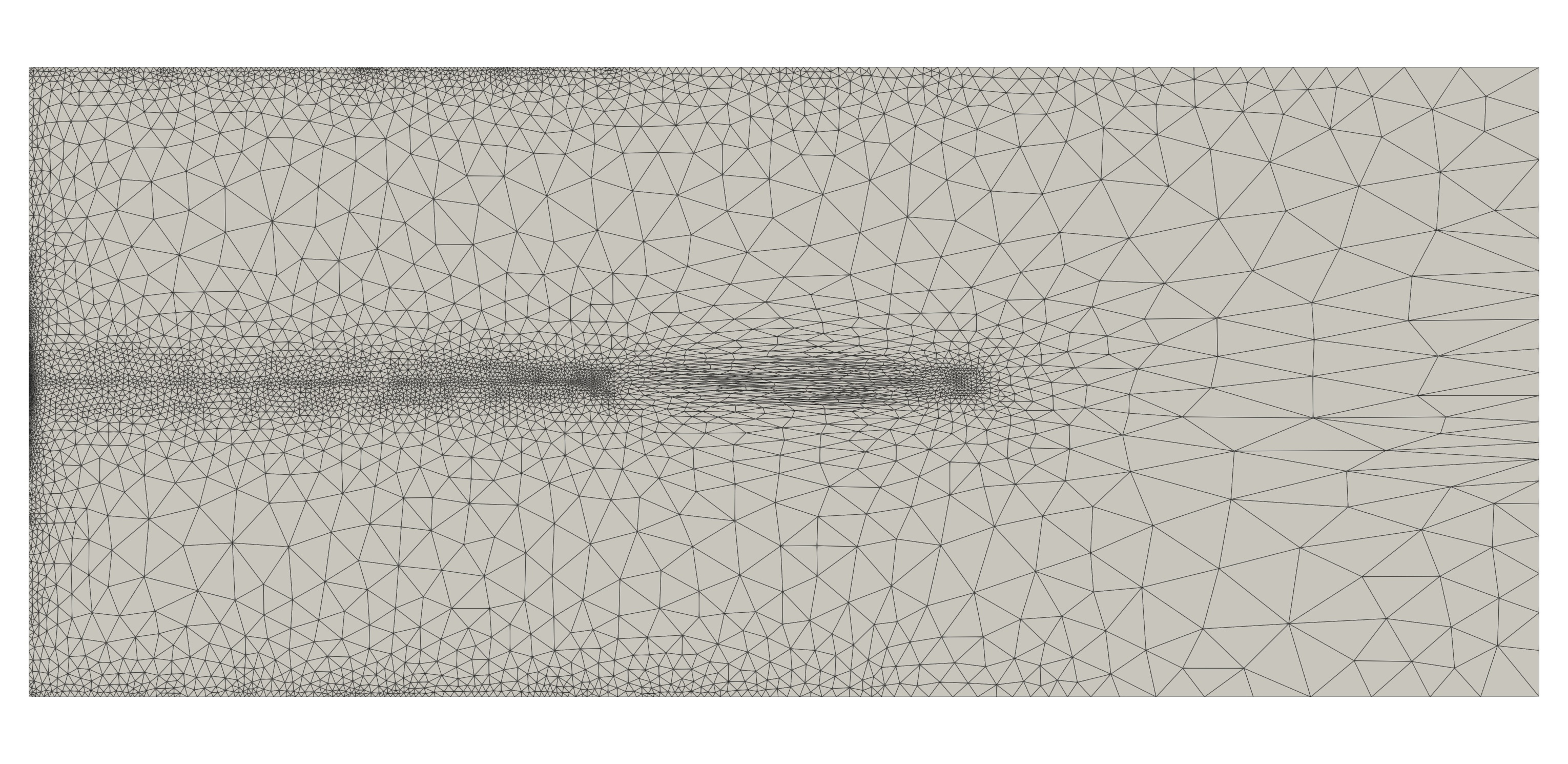}
        \caption{Aligned, goal-oriented (82,904 DoFs, 5 iterations)}
        \label{subfig:numexp:testing:meshes:aligned:GO}
    \end{subfigure}
    \begin{subfigure}{0.49\textwidth}
        \centering
        \includegraphics[width=\textwidth, trim={20 20 20 20}, clip]{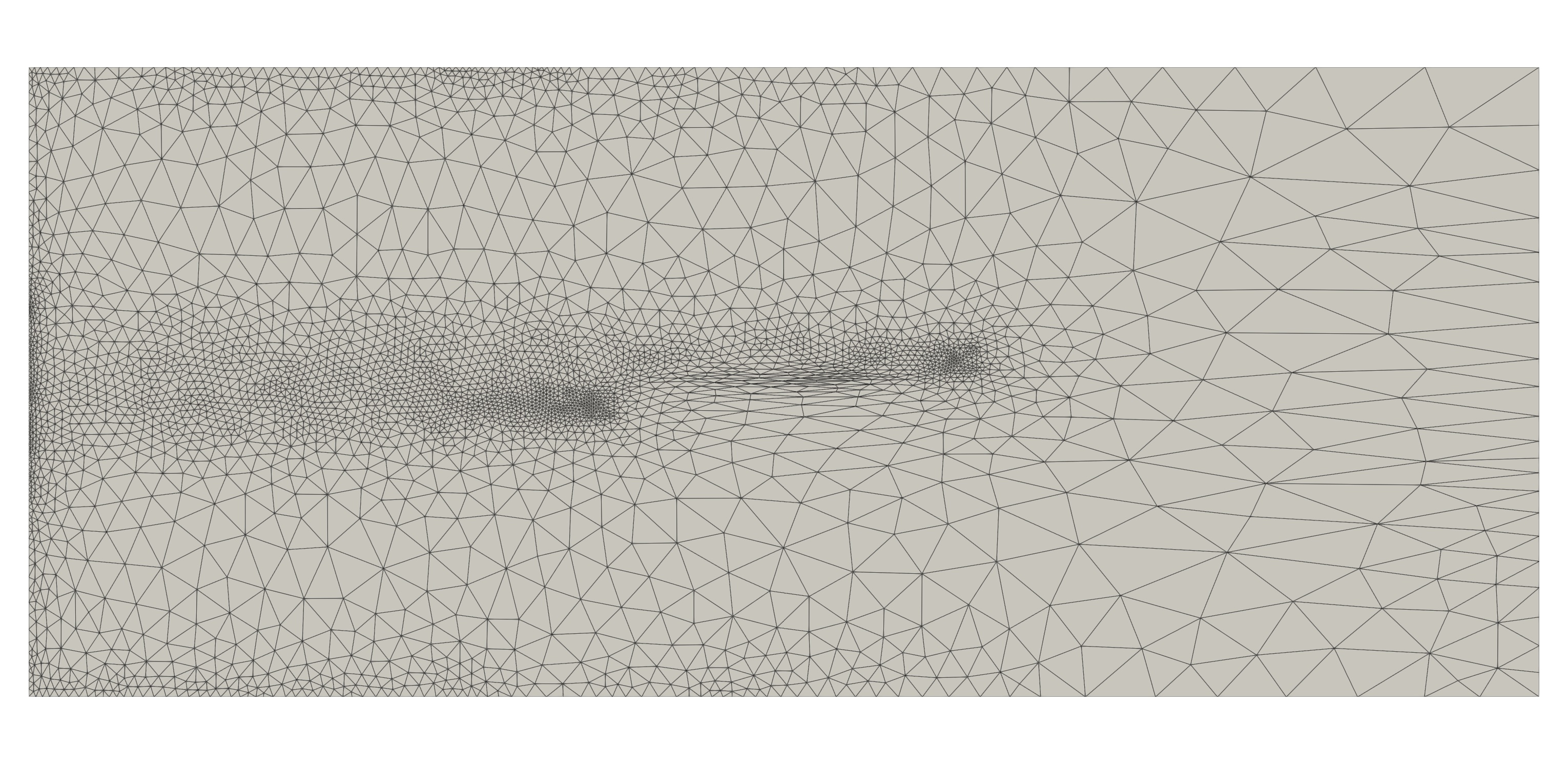}
        \caption{Offset, goal-oriented (53,617 DoFs, 4 iterations)}
        \label{subfig:numexp:testing:meshes:offset:GO}
    \end{subfigure}
    \begin{subfigure}{0.49\textwidth}
        \centering
        \includegraphics[width=\textwidth, trim={20 20 20 20}, clip]{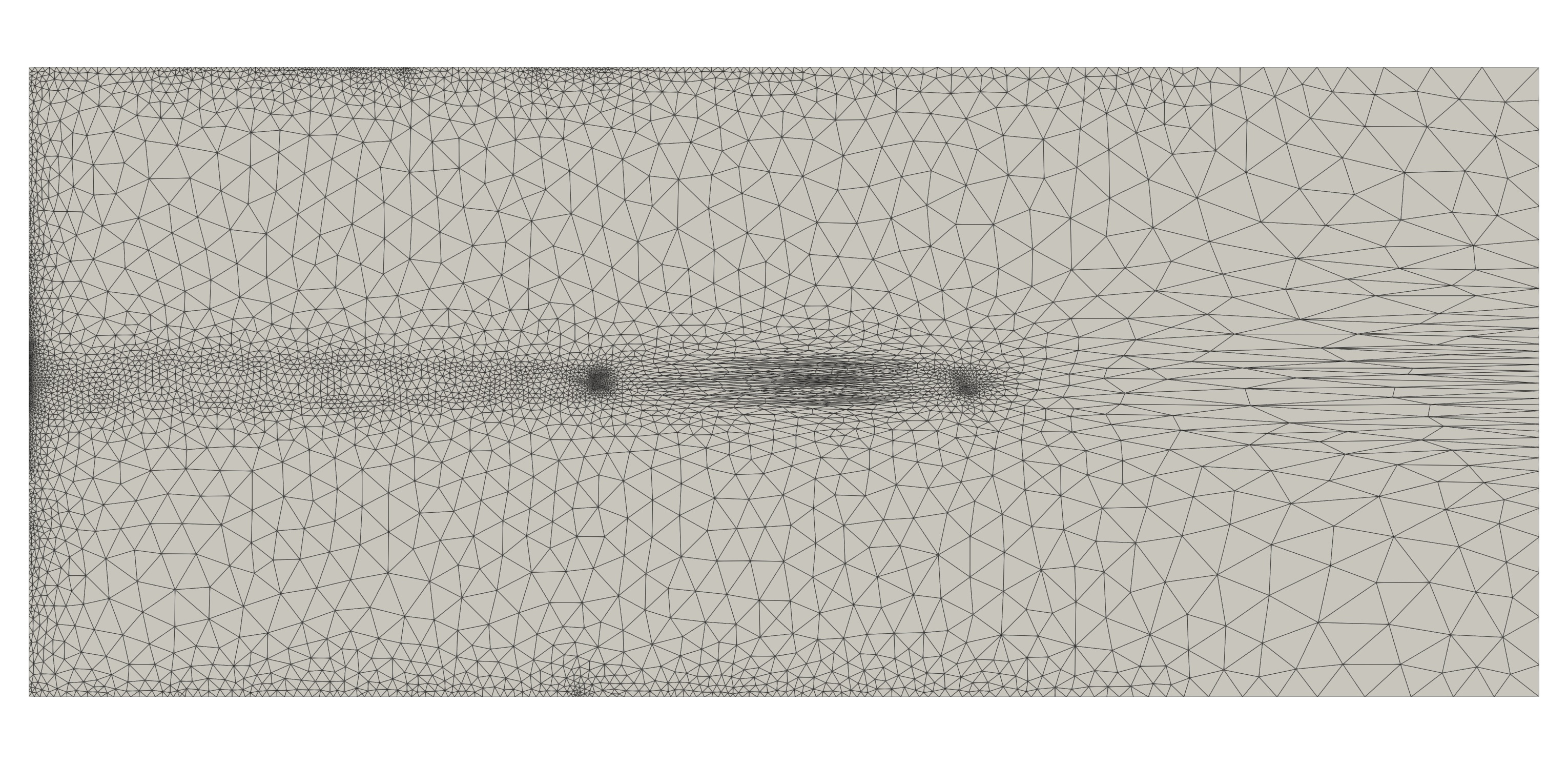}
        \caption{Aligned, data-driven (74,777 DoFs, 5 iterations)}
        \label{subfig:numexp:testing:meshes:aligned:ML}
    \end{subfigure}
    \begin{subfigure}{0.49\textwidth}
        \centering
        \includegraphics[width=\textwidth, trim={20 20 20 20}, clip]{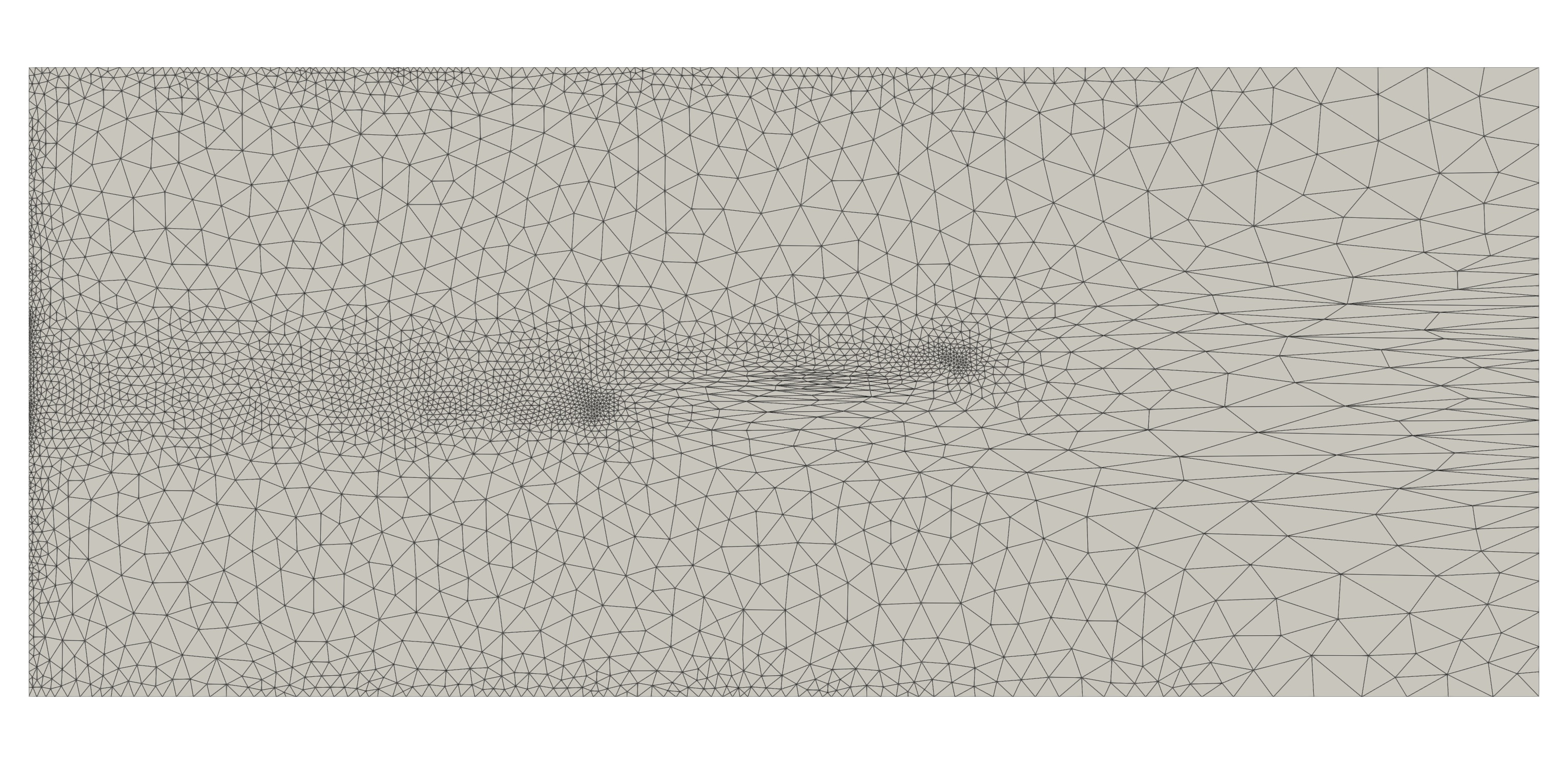}
        \caption{Offset, data-driven (50,741 DoFs, 4 iterations)}
        \label{subfig:numexp:testing:meshes:offset:ML}
    \end{subfigure}
    \caption{
        Converged adapted meshes for the `aligned' and `offset' test cases.
        Meshes generated using standard goal-oriented mesh adaptation are shown at the top, whilst the corresponding data-driven meshes are shown at the bottom.
        The number of fixed point iterations required for convergence are provided.
    }
    \label{fig:numexp:testing:meshes}
\end{figure}

The scaling of the data-driven error indicator fields does not need to be perfect to still produce high quality adaptive meshes on which the QoI can be accurately approximated.
Figure \ref{fig:numexp:testing:meshes} shows the meshes generated on convergence of the resulting fixed point iteration loops.

The top row of Figure \ref{fig:numexp:testing:meshes} shows the adapted meshes due to the standard approach.
The influence of the error indicators is clearly visible: the highest level of mesh resolution is deployed at and upstream of the turbines, there is also increased resolution along the upstream domain boundaries, and coarse resolution is used elsewhere, especially downstream.
Recall that we use a metric formulation based on the Hessian of the forward solution.
The influence of the Hessian can also be seen in the anisotropic elements used between the turbines and downstream of the turbines.
These are the regions where the curvature of the forward solution take the largest values.

The bottom row of Figure \ref{fig:numexp:testing:meshes} shows the corresponding data-driven meshes.
The similarity between the goal-oriented and data-driven adaptive methods is striking.
All of the meshing patterns observed above are also apparent for these meshes.
The main difference is that the data-driven approach leads to adapted meshes with high anisotropy in the downstream region.
This occurs because the error is not estimated to be as low as with the standard goal-oriented indicator, meaning that the Hessian's high curvature in that region is not `cancelled out' to the same extent.

\paragraph{Convergence Analysis}

\begin{figure}[t]
    \centering
    \begin{subfigure}{0.9\textwidth}
        \centering
        \includegraphics[width=\textwidth]{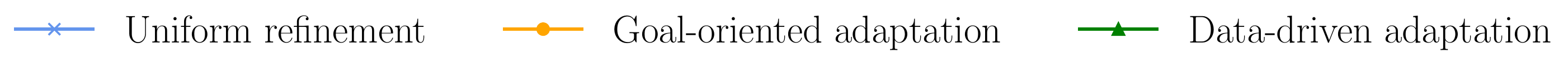}
    \end{subfigure}\\
    \begin{subfigure}{0.49\textwidth}
        \centering
        \includegraphics[width=\textwidth, trim={20 20 20 20}, clip]{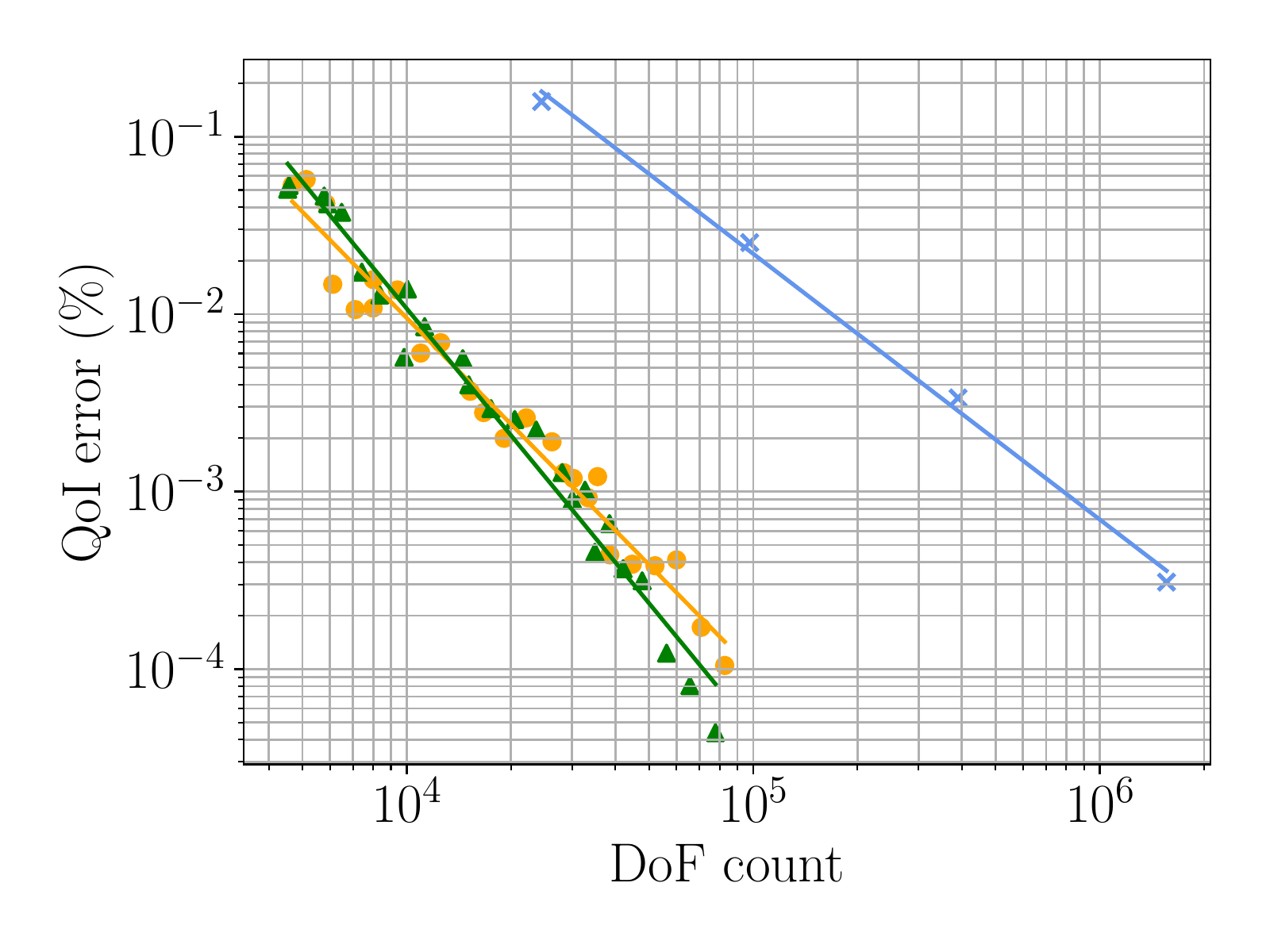}
        \caption{Aligned case}
        \label{subfig:numexp:testing:convergence:dofs:aligned}
    \end{subfigure}
    \begin{subfigure}{0.49\textwidth}
        \centering
        \includegraphics[width=\textwidth, trim={20 20 20 20}, clip]{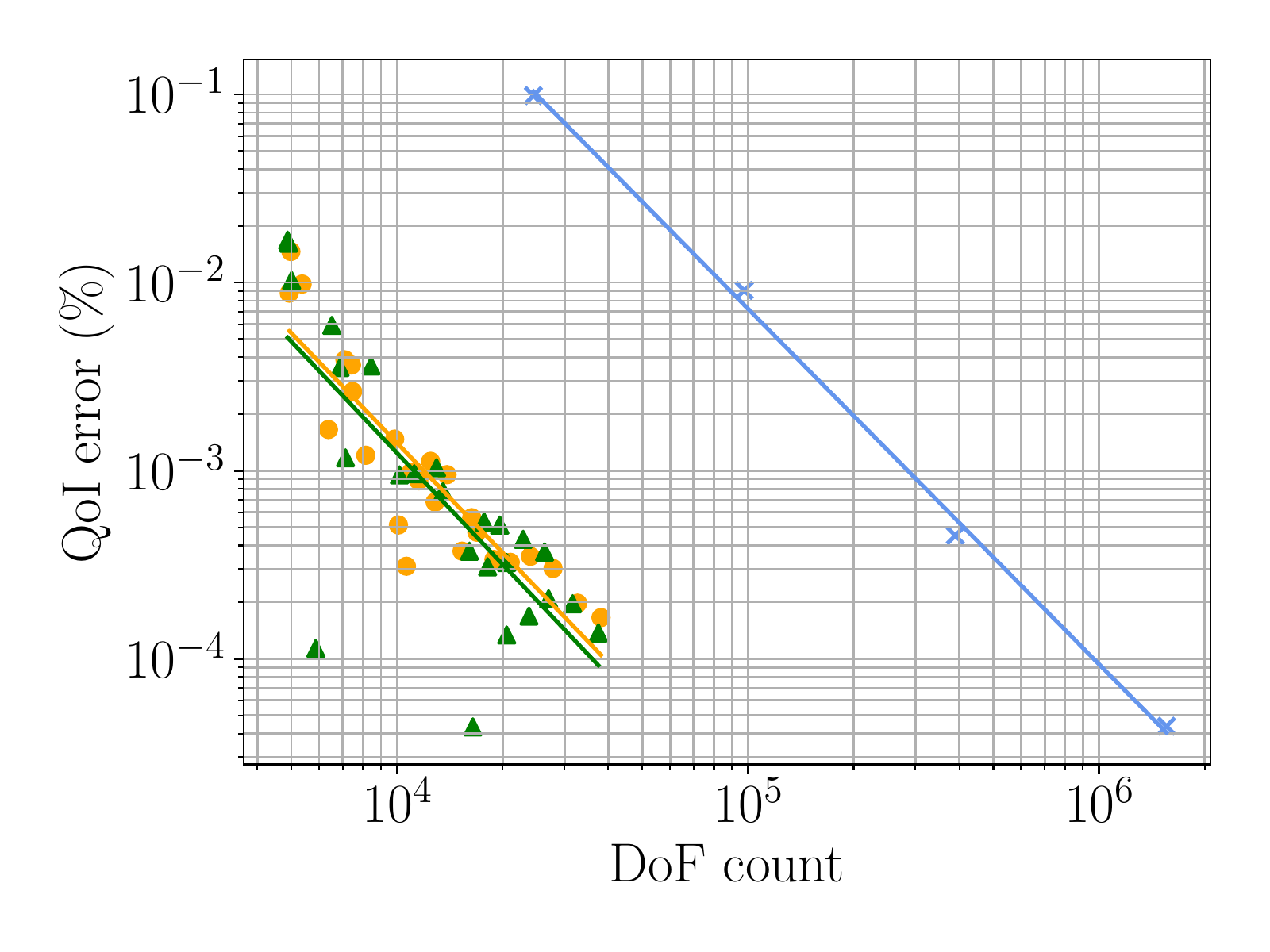}
        \caption{Offset case}
        \label{subfig:numexp:testing:convergence:dofs:offset}
    \end{subfigure}
    \begin{subfigure}{0.49\textwidth}
        \centering
        \includegraphics[width=\textwidth, trim={20 20 20 20}, clip]{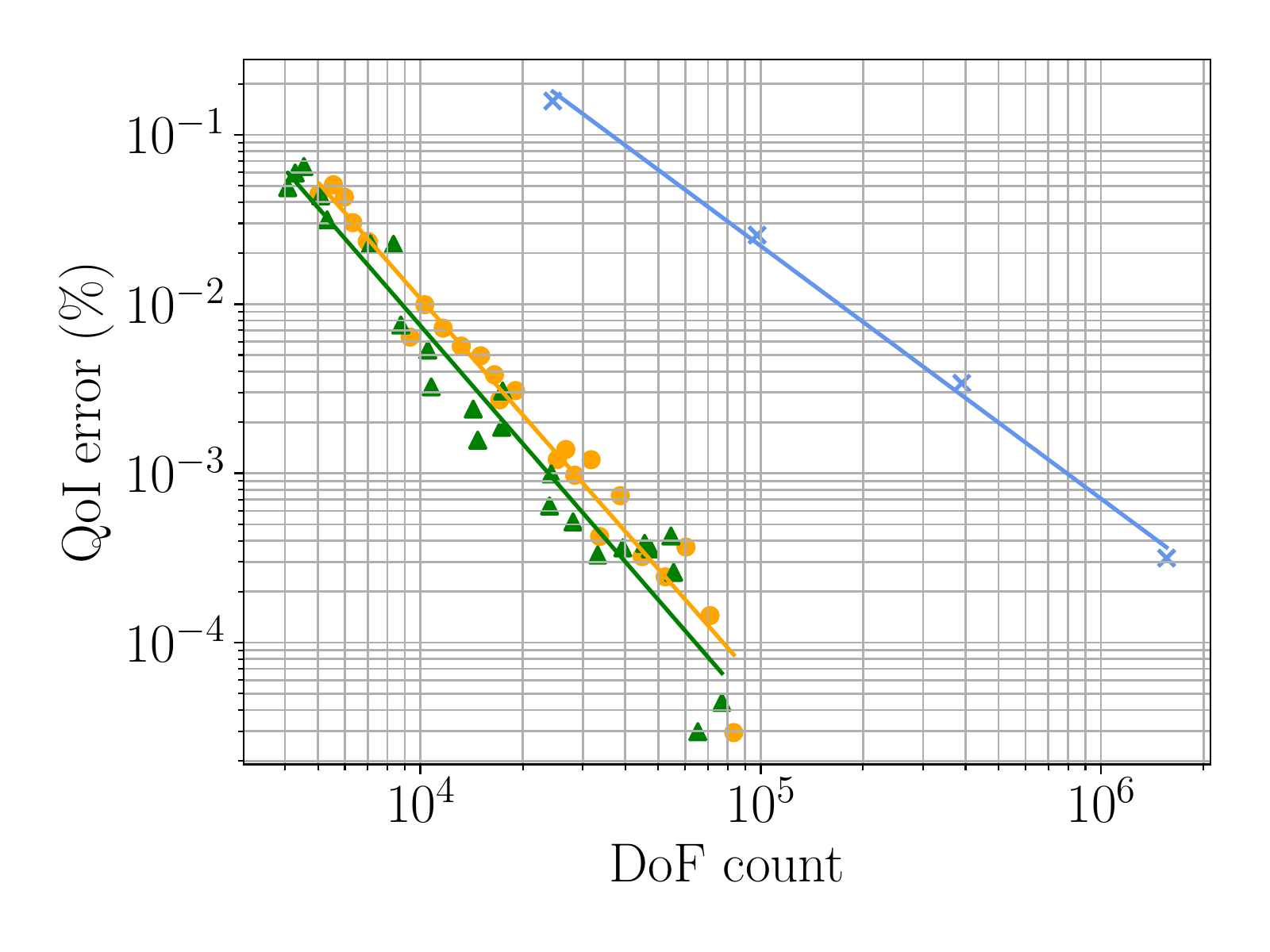}
        \caption{Reversed (aligned) case}
        \label{subfig:numexp:testing:convergence:dofs:reversed}
    \end{subfigure}
    \begin{subfigure}{0.49\textwidth}
        \centering
        \includegraphics[width=\textwidth, trim={20 20 20 20}, clip]{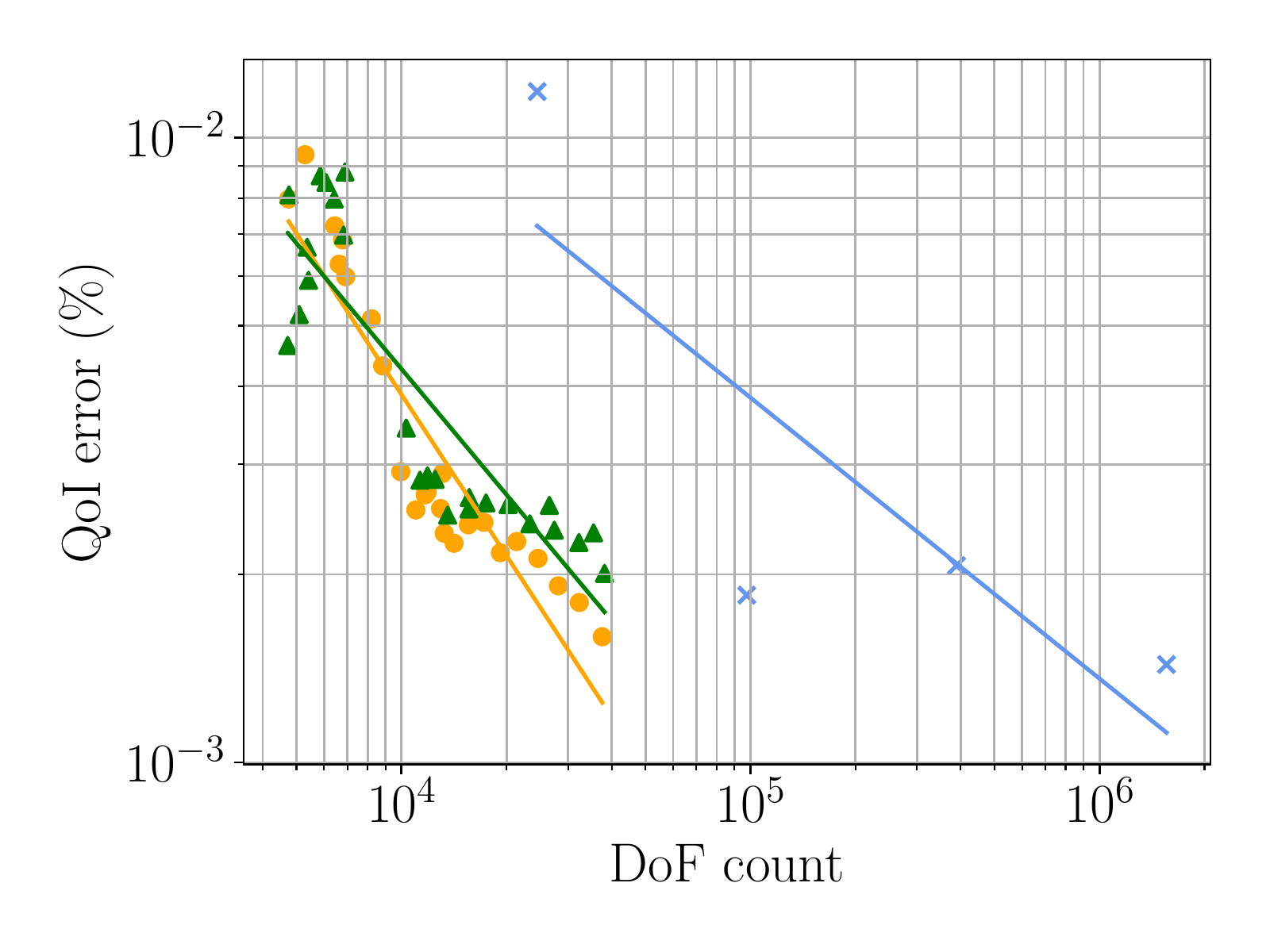}
        \caption{Trench (offset) case}
        \label{subfig:numexp:testing:convergence:dofs:trench}
    \end{subfigure}
    \caption{
        QoI error convergence curves against DoF count, for `unseen' test case configurations.
    }
    \label{fig:numexp:testing:convergence:dofs}
\end{figure}

Figure \ref{fig:numexp:testing:convergence:dofs} shows QoI error convergence curves associated with various `unseen' test cases.
The top row of plots are the ones being considered presently and the bottom row are discussed later in Subsection \ref{subsec:numexp:gen}.

In each subfigure, data points are plotted for runs of the different mesh adaptation methods with increasing overall DoF count.
We are able to request meshes with more DoFs by increasing the target metric complexity in the scaling (\ref{eq:meth:metric:scaling}).
This is independent of the error estimation step and so does not require retraining of the network.
Benchmark QoI values are obtained on high resolution fixed meshes generated by uniformly refining the initial meshes four times.
These values agree with the corresponding QoI values due to three uniform refinements to at least three significant figures.
QoI `errors' may then be calculated in a relative sense by comparison against these benchmark values.
Lines of best fit are included, to give a sense of the convergence behaviour.

In the case of uniform refinement (blue curve), Subfigures \ref{subfig:numexp:testing:convergence:dofs:aligned} and \ref{subfig:numexp:testing:convergence:dofs:offset} show that higher overall resolution leads to reduced QoI error in general and that the convergence rate is approximately constant.
Applying anisotropic goal-oriented mesh adaptation (shown in yellow), we observe roughly the same rate of convergence as with uniform refinement.
However, for a given DoF count in the mixed finite element space, the resulting QoI error values are smaller by two orders of magnitude.
The data-driven version (green) is demonstrated to be capable of matching the error convergence rate and values in both aligned and offset cases.
It is noteworthy that this is achieved, despite the fact that the data-driven error indicators do not exactly match those due to the `gold standard' method.
We reiterate that the error estimation does not need to be `perfect' in order to arrive at an effective mesh adaptation routine.

\paragraph{Computational Cost at Runtime}

\begin{figure}[t]
    \centering
    \begin{subfigure}{0.9\textwidth}
        \centering
        \includegraphics[width=\textwidth]{legend.pdf}
    \end{subfigure}
    \begin{subfigure}{0.49\textwidth}
        \centering
        \includegraphics[width=\textwidth, trim={20 20 20 20}, clip]{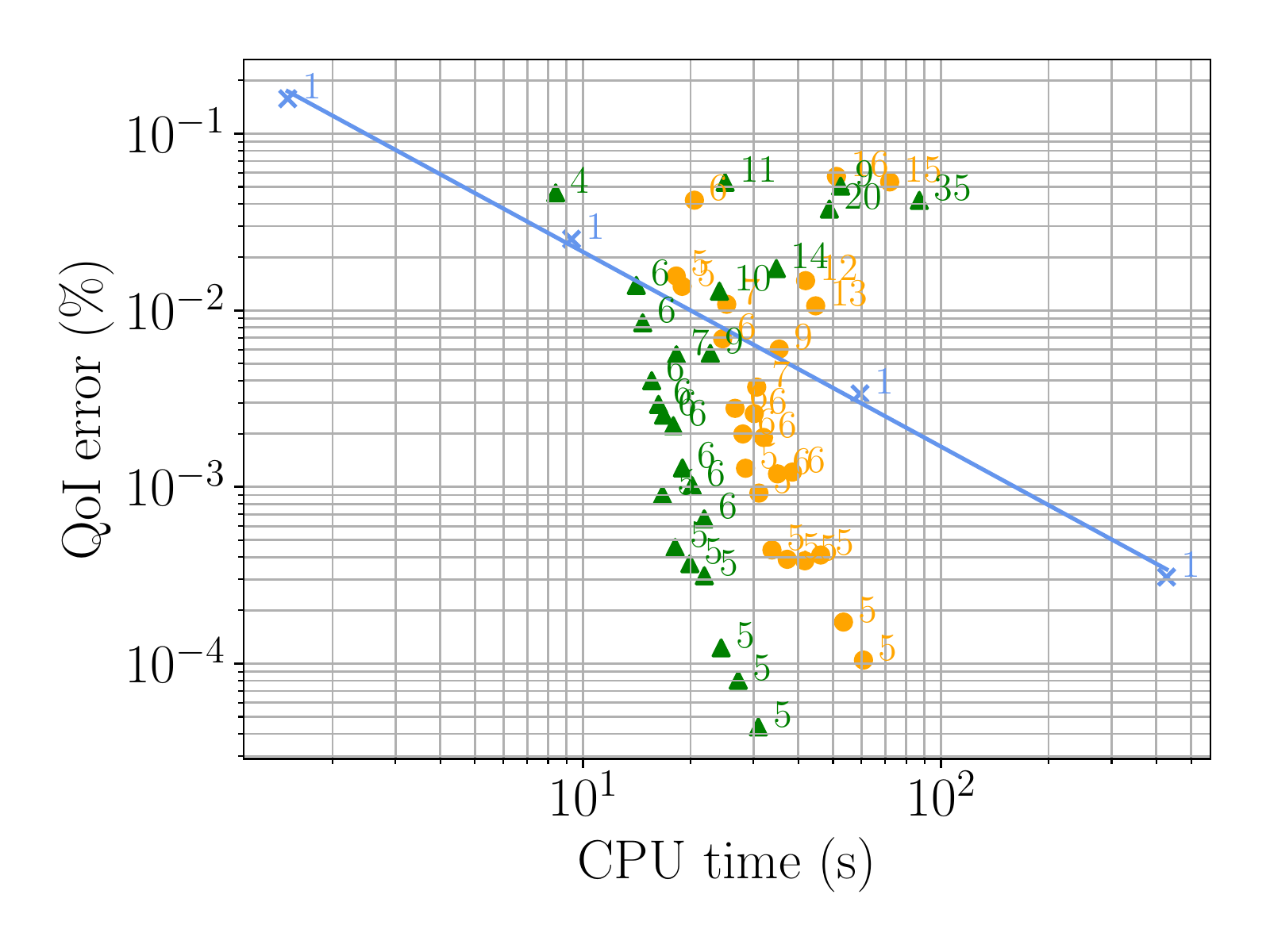}
        \caption{Aligned case}
        \label{subfig:numexp:testing:convergence:cpu:aligned}
    \end{subfigure}
    \begin{subfigure}{0.49\textwidth}
        \centering
        \includegraphics[width=\textwidth, trim={20 20 20 20}, clip]{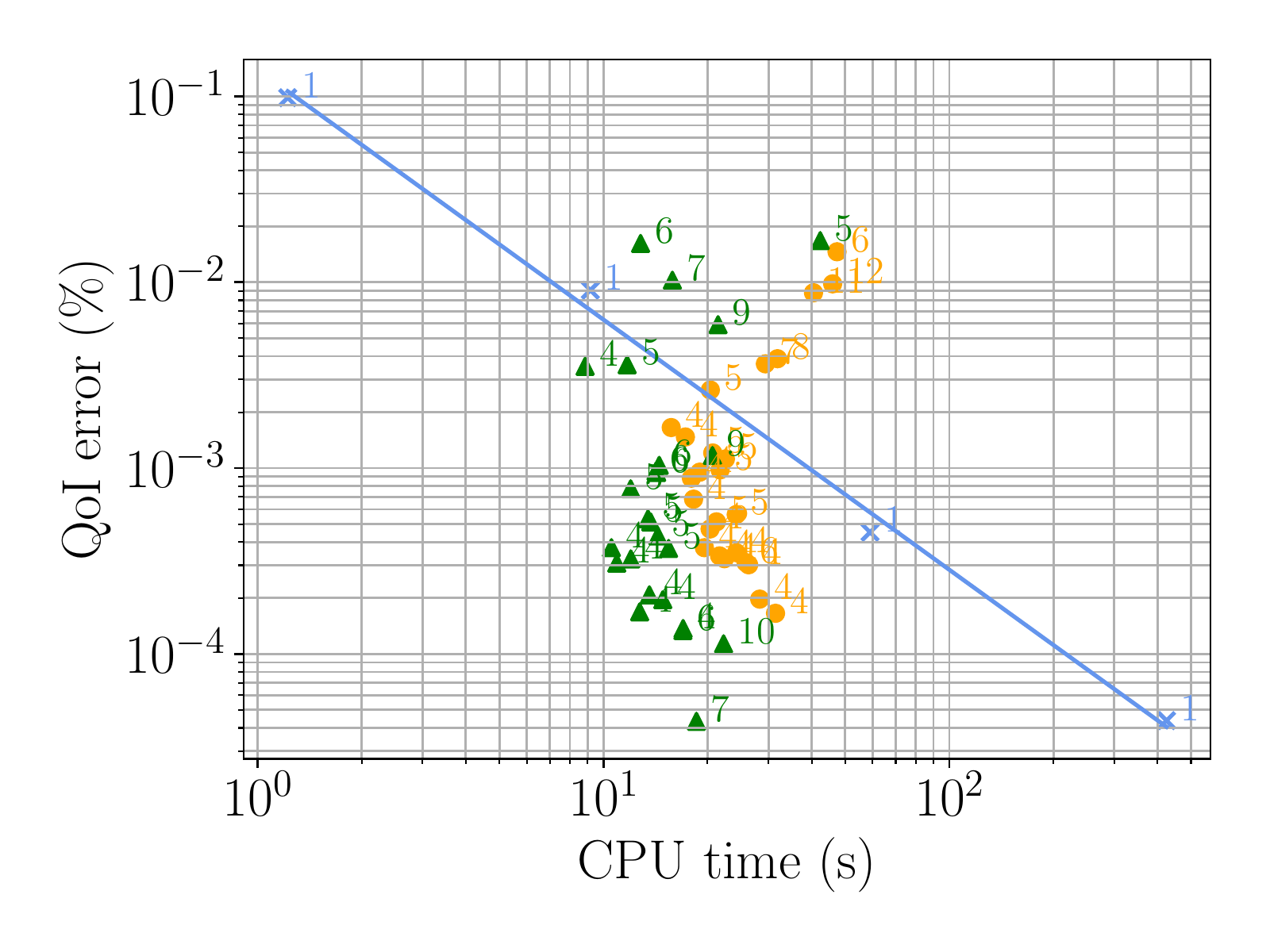}
        \caption{Offset case}
        \label{subfig:numexp:testing:convergence:cpu:offset}
    \end{subfigure}
    \caption{
        QoI error convergence curves against CPU time, for `unseen' test case configurations.
        The numbers of fixed point iterations used to generate each data point are annotated alongside the data points.
    }
    \label{fig:numexp:testing:convergence:cpu}
\end{figure}

So far we have assessed the ability of the network to emulate the goal-oriented error estimation process.
Next we assess its ability to \emph{accelerate} it, i.e.~quantify the associated reduction in computational cost.
Figure \ref{fig:numexp:testing:convergence:cpu} shows outputs from the same simulations as in Figure \ref{fig:numexp:testing:convergence:dofs}, except with CPU time being the independent variable, rather than DoF count.

Plotted against CPU time, the rate of decrease is again roughly constant for the uniform refinement approach.
This is not true for the goal-oriented and data-driven approaches.
One reason is that more fixed point iterations are typically required when the target metric complexity is low, since the adaptation routine is allowed fewer DoFs than might be preferred.
Another reason is that there are various setup costs associated with mesh adaptation that are negligible at scale, but often dominate the computational cost of small problems.
As such, the convergence curves are far from monotonic, at least for the target complexities considered.
If errors of 2\% or more are admissible, uniform refinement is the fastest approach.
As the admissible error value is reduced, however, the computational benefit of using adaptive methods increases.
For most error levels or CPU times, the data-driven approach is shown to come with an additional computational benefit over the standard goal-oriented approach.
That is, the data-driven approach is able to attain a given error level using fewer DoFs in general.

\begin{figure}[t]
    \centering
    \begin{subfigure}{0.49\textwidth}
        \centering
        \includegraphics[width=\textwidth, trim={10 5 10 5}, clip]{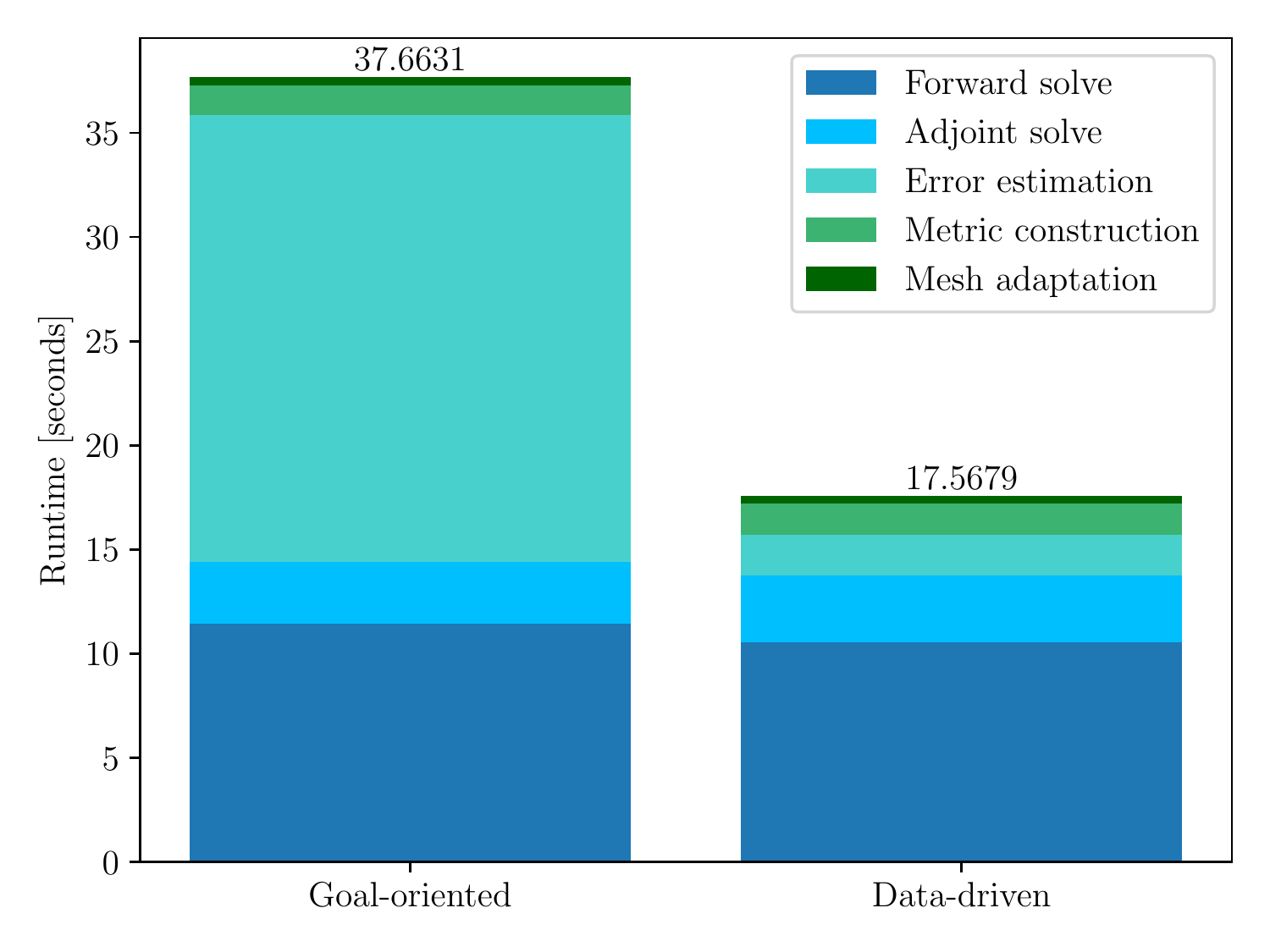} 
        \caption{Aligned (5 iterations, QoI error 0.039\%)}
        \label{subfig:numexp:testing:timing:aligned}
    \end{subfigure}
    \begin{subfigure}{0.49\textwidth}
        \centering
        \includegraphics[width=\textwidth, trim={10 5 10 5}, clip]{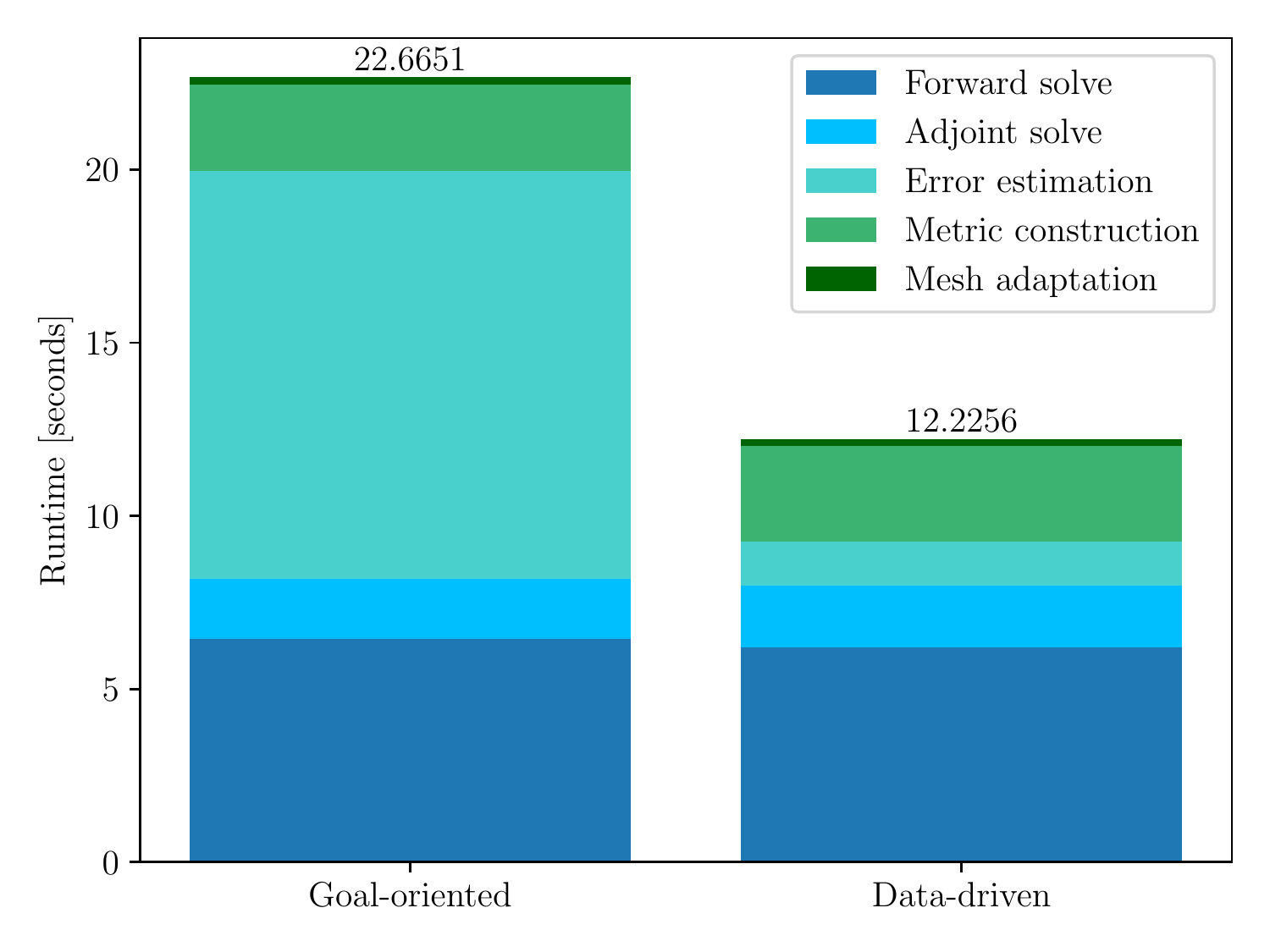} 
        \caption{Offset (4 iterations, QoI error 0.037\%)}
        \label{subfig:numexp:testing:timing:offset}
    \end{subfigure}
    \caption{
        CPU timings for each component of the mesh adaptation algorithms, for instances of the `aligned' and `offset' test cases.
        In both cases, a target complexity of 3,200 is used.
    }
    \label{fig:numexp:testing:timing}
\end{figure}

To investigate the computational cost further, we consider a `profiling' experiment, where we take simulations that give QoI estimates of comparable accuracy and then contrast the runtimes of the different components of the associated algorithms.
From the aligned test case, we take the data points generated using a target metric complexity of 3,200, both of which give rise to errors close to 0.039\%.
For the offset test case and the same target complexity, both approaches give rise to errors close to 0.037\%.
Figure \ref{fig:numexp:testing:timing} shows the timings for each component.
Applying the data-driven approach, the overall runtime can be approximately halved, compared with standard goal-oriented mesh adaptation.
We do not provide a breakdown of a fixed (uniform) mesh run that yields a similar QoI accuracy because, as may be expected, such runs consist solely of a `forward solve'.

The `forward solve' component corresponds to the solution of the nonlinear system associated with the discretised shallow water equations and the subsequent evaluation of the QoI.
The two adaptive runs converge in the same number of fixed point iterations and use the same target complexity, so the forward solve runtimes are comparable across those bars; similarly for the `adjoint solve' component.
The adjoint solves (in the base space/on the base mesh) take less time than the forward solves because adjoint equations are always linear, whereas the (forward) shallow water equations are nonlinear and their numerical solution requires multiple Newton iterations.

For the standard goal-oriented approach, the `error estimation' component includes the construction of the enriched space, the associated adjoint solve in that enriched space, assembly of the error indicator in the enriched space and finally its transfer down to the base space.
This accounts for over 56\% of the total runtime, emphasising the motivation for accelerating the error estimation process.
For the data-driven approach, the `error estimation' component includes the extraction of features for the input data and the subsequent propagation through the network.
Together, these account for less than 11\% of the data-driven runtime (less than 5\% of the original runtime), implying a significant cost reduction.
In terms of the error estimation step alone, the data-driven approach yields a 90\% saving in computational cost.
The majority of the remaining cost is due to feature extraction, since this involves some computations related to mesh element geometry, evaluating coarse-grained approximations to the error estimator and determining gradient and Hessian values at cell centroids.

The `metric construction' and `mesh adaptation' steps record the relatively small amount of time spent constructing Riemannian metrics from the error indicator data and for the mesh optmiser Mmg to use this to generate an adapted mesh, respectively.
Most of the cost associated with the metric computation is due to Hessian recovery from the forward solution.
This is done in both goal-oriented and data-driven approaches, so the cost is similar in each case.

\paragraph{Computational Cost of Training}

In the above, we show that the runtime cost of goal-oriented mesh adaptation can be reduced significantly by application of neural networks.
However, runtime improvements in efficiency are not the only important factor -- the cost of training should also be considered.
The two major costs are generating the feature and target data and then training the network on these inputs and outputs.
These are shown in Table \ref{tab:numexp:testing:timing}.
\begin{table}
    \centering
    \begin{tabular}{ccc}
        Workflow component  & Count         & Total time taken\\
        \hline
        Mesh generation     & 100 scenarios & 4.7 minutes\\
        Feature generation  & 100 scenarios & 95.5 minutes\\
        Network training    & 2,000 epochs  & 100.8 minutes\\
        \hline
                            & Overall       & 201.0 minutes
    \end{tabular}
    \caption{
        A table showing the CPU timings for each component of the neural network training.
        Temporal values are rounded to the nearest minute.
    }
    \label{tab:numexp:testing:timing}
\end{table}

Given that we train over 100 different scenarios, the cost per scenario is approximately one minute.
For a given amount of input and output data, the network training cost is then effectively a function of the number of epochs, taken here to be 2,000.
For the experiments presented in this work, the cost is approximately six seconds per epoch.
We argue that the costs are small prices to pay to double the runtime efficiency.

\subsection{Generalisation}\label{subsec:numexp:gen}

So far, we have tested the data-driven mesh adaptation approach on two `unseen' cases.
For both of these cases, the physics coefficients are well within the bounds of the parameter spaces that they are randomly sampled from, i.e.~$\nu=0.5\in[0.1,1]$, $b=40\in[10,100]$ and $u_{\mathrm{inflow}}=5\in[0.5,6]$.
In order to test the approach more thoroughly, we consider some additional test cases whose problem specifications have key differences from the training cases.

\subsubsection{Reversed Flow Direction}\label{subsubsec:numexp:gen:rev}

All cases considered thus far have flow in the positive $x$-direction.
Consider an identical setup to the `aligned' test case, but with the inflow and outflow boundaries switched and the sign of the inflow velocity reversed: $u_{\mathrm{inflow}}=-5\,\mathrm{m\,s}^{-1}$.

\begin{figure}[t]
    \centering
    \begin{subfigure}{0.49\textwidth}
        \centering
        \includegraphics[width=\textwidth, trim={20 20 20 20}, clip]{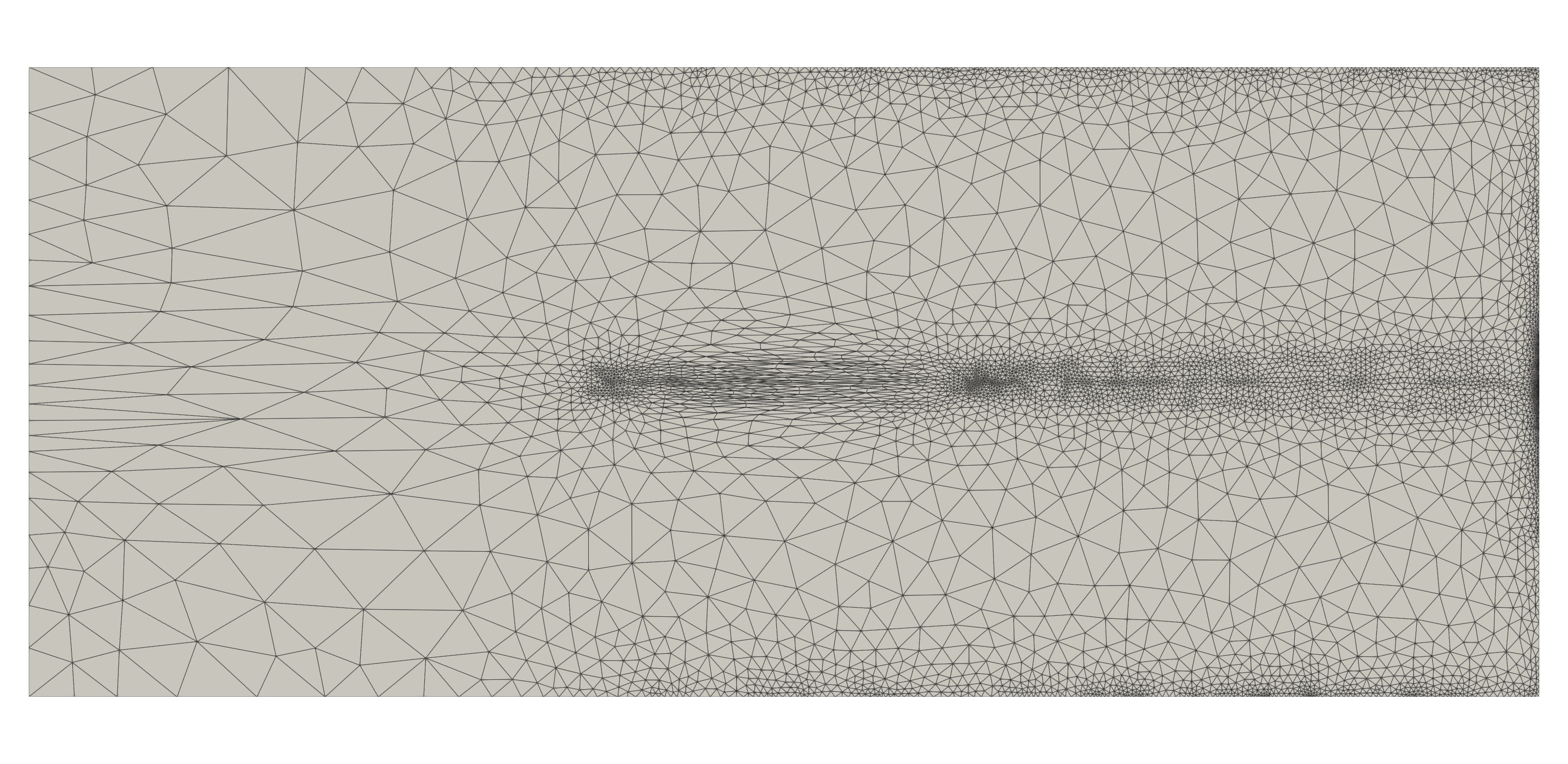}
        \caption{Reversed, goal-oriented (84,281 DoFs)}
        \label{subfig:numexp:gen:rev:meshes:GO}
    \end{subfigure}
    \begin{subfigure}{0.49\textwidth}
        \centering
        \includegraphics[width=\textwidth, trim={20 20 20 20}, clip]{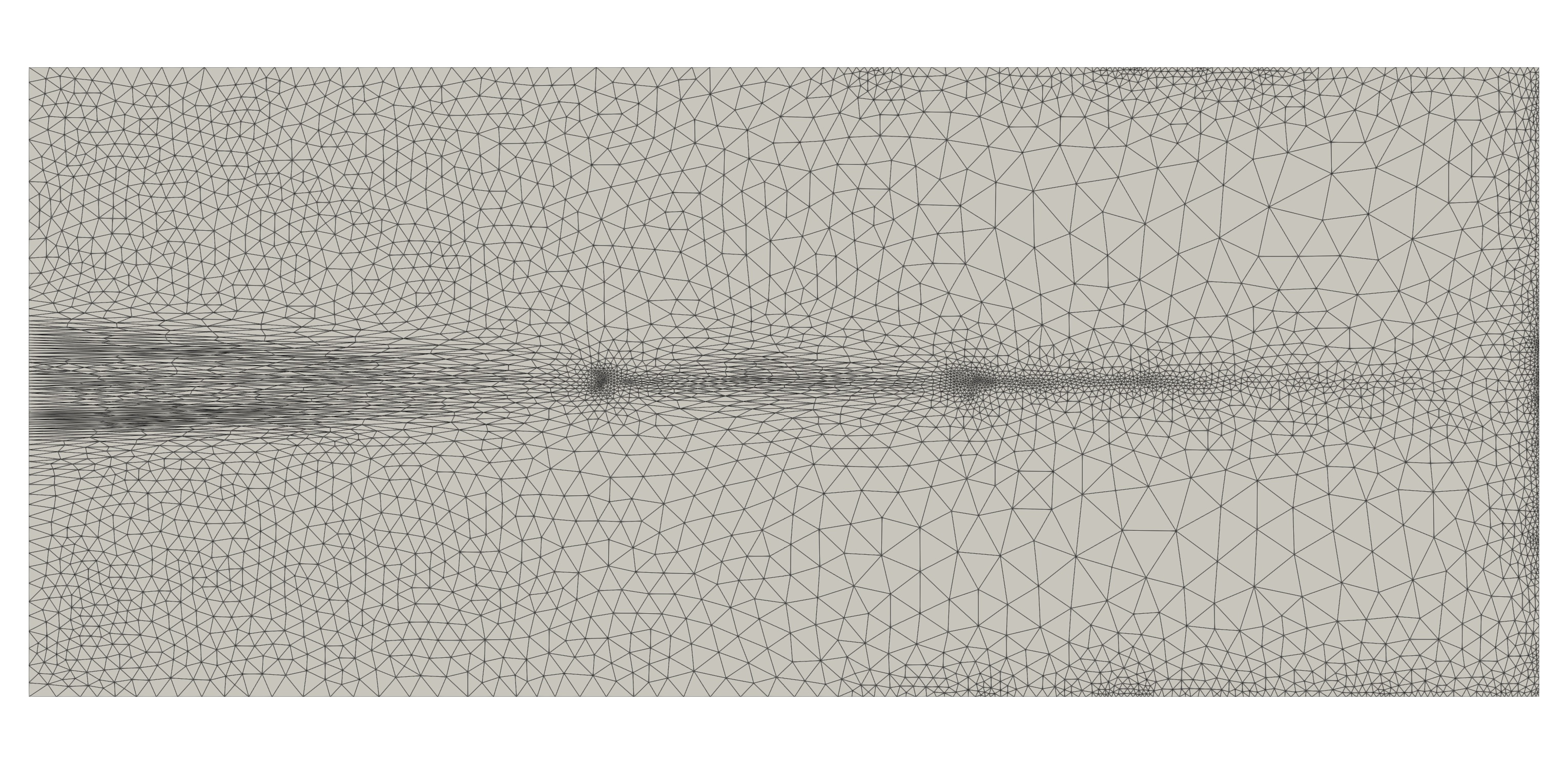}
        \caption{Reversed, data-driven (75,131 DoFs)} 
        \label{subfig:numexp:gen:rev:meshes:ML}
    \end{subfigure}
    \begin{subfigure}{0.49\textwidth}
        \centering
        \scalebox{-1}[1]{\includegraphics[width=\textwidth, trim={20 20 20 20}, clip]{aligned_mesh_GO_0004.jpeg}}
        \caption{Reflected, goal-oriented (82,904 DoFs)}
        \label{subfig:numexp:gen:ref:meshes:GO}
    \end{subfigure}
    \begin{subfigure}{0.49\textwidth}
        \centering
        \scalebox{-1}[1]{\includegraphics[width=\textwidth, trim={20 20 20 20}, clip]{aligned_mesh_ML_0004.jpeg}}
        \caption{Reflected, data-driven (74,777 DoFs)}
        \label{subfig:numexp:gen:ref:meshes:ML}
    \end{subfigure}
    \caption{
        Top row: Converged adapted meshes for a reversed flow version of the `aligned' test case.
        The mesh generated using standard goal-oriented mesh adaptation is shown on the left, whilst the corresponding data-driven mesh is shown on the right.
        Both `reversed' cases are run for four fixed point iterations, for consistency with the original runs.
        Bottom row: Reflected versions of the corresponding original results from Subfigures \ref{subfig:numexp:testing:meshes:aligned:GO} and \ref{subfig:numexp:testing:meshes:aligned:ML}.
    }
    \label{fig:numexp:gen:rev:meshes}
\end{figure}

The top row of Figure \ref{fig:numexp:gen:rev:meshes} shows the converged adaptive meshes for both the standard goal-oriented and data-driven approaches.
These plots exhibit the same meshing patterns as observed in Subfigures \ref{subfig:numexp:testing:meshes:aligned:GO} and \ref{subfig:numexp:testing:meshes:aligned:ML}, but reflected from left to right.
Reflected versions of the plots from those subfigures are shown in the bottom row, for comparison.
The standard goal-oriented meshes take a very similar form; it is difficult to pick out any notable differences.
The data-driven meshes are also somewhat comparable, although the band of resolution upstream (i.e.~to the right) of the turbines is less apparent in the `reversed' case.
In addition, errors appear to be over-estimated in the downstream region (i.e.~to the left), making the Hessian's contribution to the metric even more prominent.

Recall the convergence plot for the `aligned' test case in Subfigure \ref{subfig:numexp:testing:convergence:dofs:aligned}.
The equivalent plot for the `reversed' test case is shown in Subfigure \ref{subfig:numexp:testing:convergence:dofs:reversed}.
Whilst the two plots are not identical, they are in good agreement and the improvement on uniform refinement is clear.
As such, whilst the data-driven approach deploys mesh resolution in a slightly different way in the `reversed' test case, the overall accuracy is still of a high standard.

\subsubsection{Variable Bathymetry}\label{subsubsec:numexp:gen:trench}

Next, we consider again the `offset' setup but modify the physical parameters so that they are outside of the bounds of those used in the training data.
Moreover, we use a spatially varying bathymetry field,
\begin{equation}
    b\left(\widetilde y\right)=160+40\,\widetilde y\,\left(1-\widetilde y\right),
\end{equation}
where $\widetilde y$ is a linearly rescaled $y$-coordinate so that $\widetilde y=0$ is the bottom boundary and $\widetilde y=1$ is the top.
This means $b(\widetilde y)\not\in[10,100],\:\forall\widetilde y$.
The bathymetry forms a parabolic valley, so we refer to this as the `trench' test case.
Note that a consequence of this variable bathymetry is a more complex flow field.
The bathymetry field is represented in $\mathbb P0$ space, to coincide with how features are extracted cell-wise.
The other physical parameters are also set to be well outside of the ranges of values seen previously: $\nu=2.0\not\in[0.1,1]$; $u_{\mathrm{inflow}}=10\not\in[0.5,6]$.

\begin{figure}[t]
    \centering
    \begin{subfigure}{0.49\textwidth}
        \centering
        \includegraphics[width=\textwidth, trim={20 20 20 20}, clip]{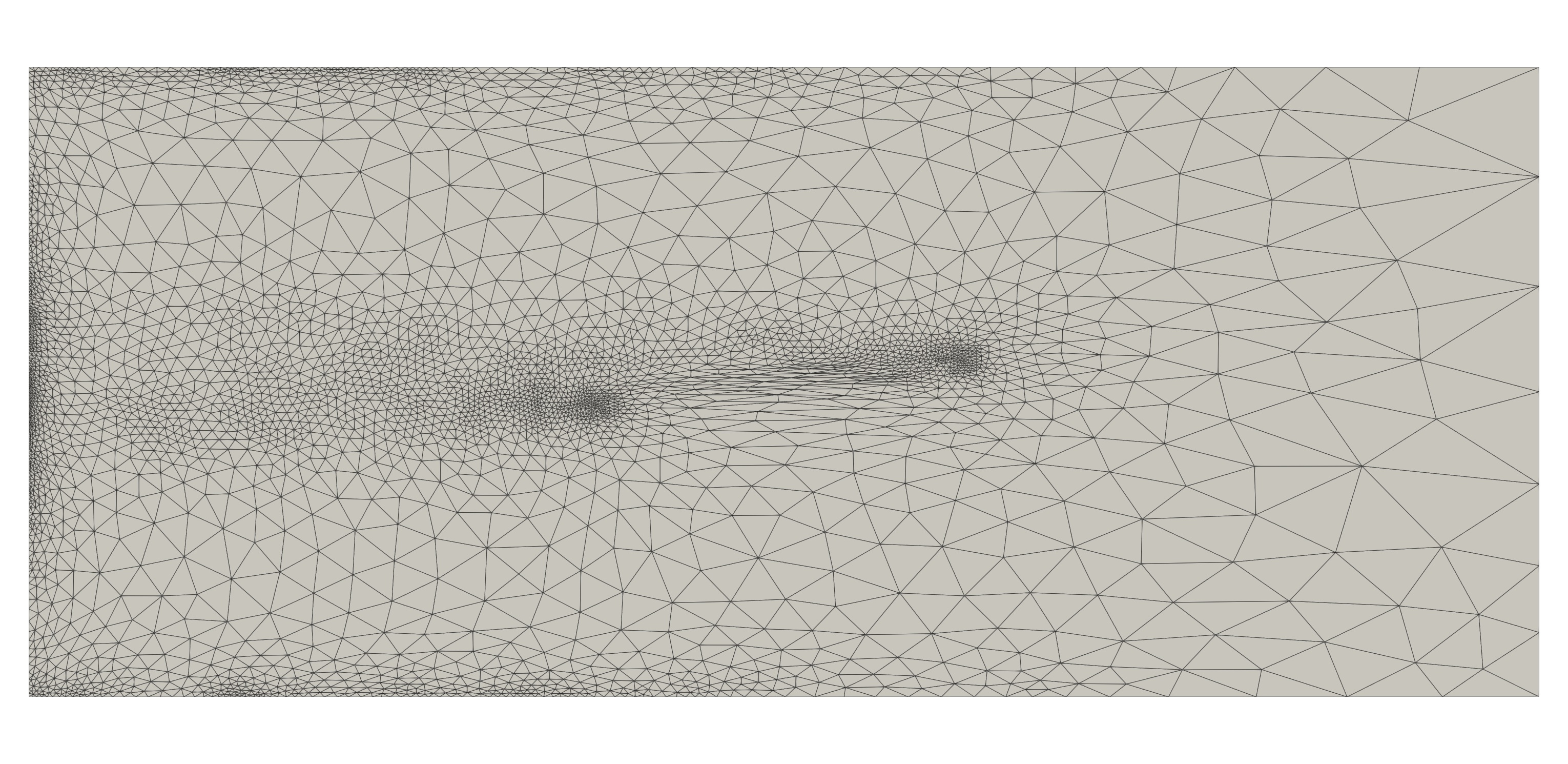}
        \caption{Goal-oriented (54,743 DoFs, 4 iterations)}
        \label{subfig:numexp:gen:trench:meshes:GO}
    \end{subfigure}
    \begin{subfigure}{0.49\textwidth}
        \centering
        \includegraphics[width=\textwidth, trim={20 20 20 20}, clip]{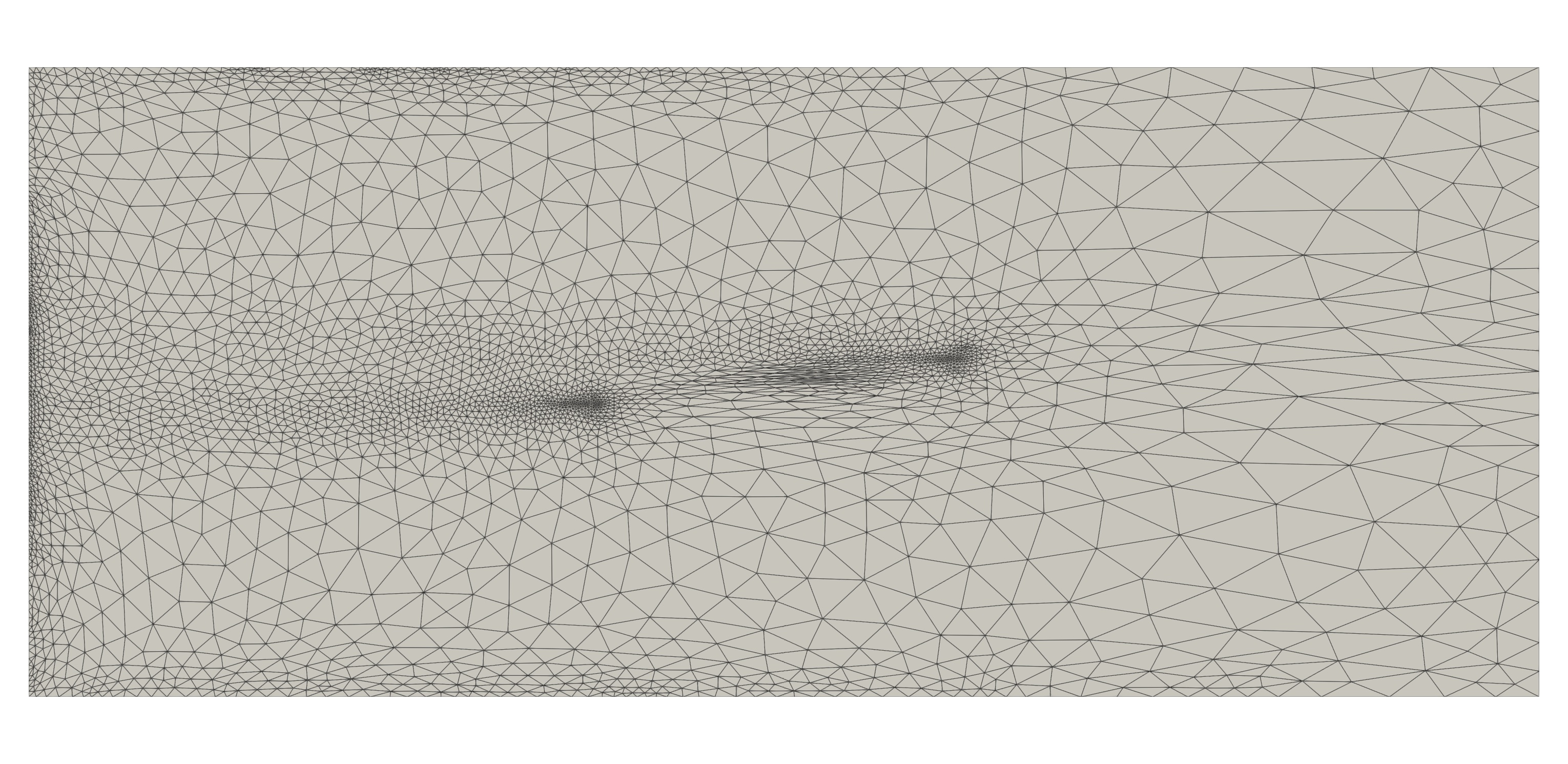}
        \caption{Data-driven (50,792 DoFs, 4 iterations)}
        \label{subfig:numexp:gen:trench:meshes:ML}
    \end{subfigure}
    \caption{
        Converged adapted meshes for the `offset' setup with physical parameters outside the bounds of the training data, including the use of a variable bathymetry field.
        The mesh generated using standard goal-oriented mesh adaptation is shown on the left, whilst the corresponding data-driven mesh is shown on the right.
        The number of fixed point iterations required for convergence are provided.
    }
    \label{fig:numexp:gen:trench:meshes}
\end{figure}

Figure \ref{fig:numexp:gen:trench:meshes} shows the resulting goal-oriented and data-driven meshes, for which QoI convergence is achieved after four fixed point iterations.
Compared with the original `offset' test case, the relatively high viscosity value exacerbates the spread of mesh resolution upstream of the turbines, as expected.
The data-driven adapted mesh is in good agreement with the one generated using standard goal-oriented adaptation.
The only noticeable difference is that the standard method puts slightly more mesh resolution at the turbines themselves.

Convergence analysis for the `trench' test case is shown in Subfigure \ref{subfig:numexp:testing:convergence:dofs:trench}, with the original `offset' results shown in Subfigure \ref{subfig:numexp:testing:convergence:dofs:offset}.
The trend of the uniform refinement curve is not as clear for this test case as for the previous ones.
Nonetheless, assuming again that the highest resolution uniform refinement run gives a good approximation to the true QoI value, we find that both the standard goal-oriented and the data-driven approaches are able to achieve comparable accuracy using around an order of magnitude fewer DoFs.

\section{Conclusion and Outlook}\label{sec:conc}

To the best of the authors' knowledge, this paper presents the first framework for using neural networks to accelerate specifically the error estimation step of a goal-oriented mesh adaptation pipeline.
The element-based approach taken in this paper helps to improve the generalisability of the neural network approach, since the global structure of the problem at hand is not of significance.

A feed-forward fully connected neural network is trained on 100 randomly generated tidal turbine modelling test cases, each with varying numbers of turbines and uniformly distributed physical parameters.
The network is then tested on two established test cases, whose associated parameters fall within the bounds of the training data.
For a given level of accuracy in the QoI approximation, the data-driven approach is found to match the standard goal-oriented approach in the sense that it is also able to achieve that accuracy level using an order of magnitude fewer DoFs than uniform refinement.
Further, the data-driven approach is shown to be capable of halving the total runtime of standard goal-oriented mesh adaptation (including the forward flow solve step), by completely avoiding the computationally expensive enrichment step.
The remaining runtime for the data-driven error estimation step (comprised of feature extraction and network propagation) is 90\% lower than for the standard approach.

Regarding inputs to the neural network, we propose here a feature set that takes into account the local element geometry, physical parameters and the variation of the forward and adjoint solution fields over.
Derivative and curvature information are included as well as values, in order to fully capture the solution fields in each element.
It is plausible that the network would further benefit from having access to information from the surrounding elements, too, giving a `patch-wise' approach.
Perhaps such an approach would retain generalisability, whilst improving its estimation capability by accessing more information on error distribution.
Note in particular that patch-wise information is especially relevant to DG discretisations, where field values tend to differ on each side of inter-element boundaries.

In addition to testing the data-driven mesh adaptation approach on existing test cases for goal-oriented mesh adaptation that were not used for training, but which have parameters within the ranges ``seen'' during training, we perform generalisation experiments, designed to test the extent to which ``unseen'' parameter values and spatially varying bathymetry fields can be handled.
Differences in meshing patterns are observed, but the QoI error convergence properties are still desirable.
Even with these extensions, however, the applicability of the data-driven approach (as presented) remains limited within the scope of all scientifically relevant shallow water modelling problems.
In terms of steady-state problems, it would be advantageous to be able to apply the data-driven approach to scenarios with different boundary condition formulations, spatially varying viscosity fields and more complex geometries, including non-convex domains with ``holes'' or ``islands''.
Preliminary investigations undergone as part of this work found that further training data must be considered if the data-driven approach is to be extended significantly in any of these directions.

The shallow water equations are often solved in their time-dependent formulation rather than to steady-state.
As such, a major piece of future work is to train a network to perform goal-oriented error estimation for time-dependent problems, such as the tidally reversing array example considered in \cite{WA22}.
Conceptually, this would be very similar to the present work, mapping feature data to error indicators.
Perhaps the training could be done on a timestep-by-timestep basis, in the same way that we work on an element-by-element basis in space.
The main difference would be that the error indicator in question is based on the residual of a time-dependent problem, as defined by the time integrator being used.
In particular, it would include an additional time derivative term.

Building upon the advancements proposed above, the ultimate goal of this work is to use neural networks to accelerate goal-oriented error estimation and mesh adaptation in the context of realistic, multi-scale coastal ocean modelling problems, and indeed a wider range of fluid dynamical and PDE based problems that can benefit from the use of goal-oriented mesh adaptation.
In such cases, the use of variable resolution is imperative to capture the multiple scales without extortionately high mesh complexities.
Goal-oriented mesh adaptation provides a way to reduce the mesh complexity, but a data-driven approach could additionally reduce runtimes.

\appendix

\section{Derivative Recovery}\label{sec:recovery}

The Riemannian metric construction used in this work makes use of the Hessians of the shallow water solution component fields.
Since the velocity components are approximated in $\mathbb P1_{DG}$ space, they are not twice continuously differentiable.
As such, we require the facility to recover derivative information from a lower order field.
We do this using the Cl\'ement interpolant \cite{Cle75}, which provides a means of transferring data between $\mathbb P0$ spaces and $\mathbb P1$ spaces via weighted averaging over neighbouring cells.

Suppose $u_h$ is a (scalar-valued) $\mathbb P1_{DG}$ field defined on some mesh, which approximates a smooth field $u$.
Its finite element gradient, $\nabla u_h$, is then a (vector-valued) $\mathbb P0$ field.
Applying the Cl\'ement interpolant, we obtain a (vector-valued) $\mathbb P1\subseteq\mathbb P1_{DG}$ field, $\mathbf g_h$, which approximates the gradient $\nabla u$.
Applying the same logic again, the finite element gradient $\nabla\mathbf g_h$ is a (matrix-valued) $\mathbb P0$ field and its Cl\'ement interpolant $\underline{\mathbf H_h}$ is a (matrix-valued) $\mathbb P1$ field, which approximates the Hessian $\nabla\nabla u$.

Given that velocity is vector-valued, recovering its Hessian essentially means applying the above procedure to each of its components.

\section{Software}\label{sec:software}

The shallow water equations are solved using the \emph{Thetis} coastal ocean model \cite{Thetis}, which is built on top of the \emph{Firedrake} \cite{Firedrake} finite element library.
Firedrake is a Python-based package, which automatically generates C code and uses \emph{PETSc} \cite{petsc-web-page,petsc-user-ref} to provide its unstructured mesh concept and to solve the linear and nonlinear systems which underpin finite element problems.
Metric-based mesh adaptation is achieved using \emph{Mmg} \cite{Mmg}, the coupling of which with PETSc's mesh representation was documented in \cite{WK22}.

The software versions used to generate the numerical experiments in this paper are archived on Zenodo at \cite{thetis_zenodo} and \cite{firedrake_zenodo}.
Simulation code is archived at \cite{code_zenodo}.

\subsection*{\bf{}\textup Acknowledgements}

\noindent The authors would like to thank members of the Applied Modelling and Computation Group (AMCG) at Imperial College London for the many interesting discussions regarding this work, in particular Stephan C. Kramer and James Tlhomole.

This work was funded in part under the embedded CSE programme of the ARCHER2 UK National Supercomputing Service (\texttt{http://www.archer2.ac.uk}).
Support is also acknowledged from EPSRC under grant EP/R029423/1.

\subsection*{\bf{}CRediT Statement}

\textbf{Joseph G. Wallwork}: Conceptualisation, Methodology, Formal Analysis, Investigation, Writing - Original Draft, Writing -- Review \& Editing, Software, Visualisation.
\textbf{Jingyi Lu}: Methodology, Investigation, Writing -- Review \& Editing.
\textbf{Mingrui Zhang}: Methodology, Writing - Review \& Editing.
\textbf{Matthew D. Piggott}: Methodology, Writing -- Review \& Editing, Supervision, Funding Acquisition.

\nocite{*}
\bibliographystyle{elsarticle-num} 
\bibliography{references.bib}
\end{document}